%% file: main.tex
\documentclass[10pt,twocolumn,letterpaper]{article}

\usepackage{wacv}
\usepackage{times}
\usepackage{epsfig}
\usepackage{graphicx}
\usepackage{amsmath}
\usepackage{amssymb}
\usepackage{booktabs}
\usepackage{amssymb}
\usepackage{pifont}
\usepackage{float}
\usepackage{subcaption}
\usepackage{lipsum}
\newcommand{\cmark}{\ding{51}}%
\newcommand{\xmark}{\ding{55}}%
\usepackage[titletoc]{appendix}
\usepackage{xcolor}         

\newcommand{\nvspace}{\vspace{-0.2cm}}
\newcommand{\tvspace}{\vspace{-0.2cm}}

\def\wacvPaperID{1877} 

\wacvapplicationstrack 

\ifwacvfinal
\usepackage[pagebackref=true,breaklinks=true,colorlinks,bookmarks=false]{hyperref}
\else
\usepackage[pagebackref=true,breaklinks=true,colorlinks,bookmarks=false]{hyperref}
\fi

\begin{document}

\title{VL-Taboo: An Analysis of Attribute-based Zero-shot Capabilities of Vision-Language Models}


\author{Felix Vogel\\
Goethe University Frankfurt\\
{\tt\small s3618056@stud.uni-frankfurt.de}
\and
Nina Shvetsova\\
Goethe University Frankfurt\\
{\tt\small shvetsov@uni-frankfurt.de}
\and
Leonid Karlinsky\\
MIT-IBM Watson AI Lab\\
{\tt\small leonidka@ibm.com}
\and
Hilde Kuehne\\
Goethe University Frankfurt\\
MIT-IBM Watson AI Lab\\
{\tt\small kuehne@uni-frankfurt.de}
}

\maketitle
\thispagestyle{empty}

\input{0_abstract.tex}

\input{1_introduction.tex}
\input{2_relatedwork.tex}

\input{3_method.tex}
\input{4_datasets.tex}

\input{5_experiments.tex}

\input{6_conclusion.tex}

\pagebreak

{\small
\bibliographystyle{ieee_fullname}
\bibliography{egbib}
}
\renewcommand{\appendixpagename}{Supplementary}
\pagebreak
\clearpage
\appendixpage
\begin{appendices}
\input{supplement.tex}

\end{appendices}

\end{document}

%% file: 0_abstract.tex
\begin{abstract}
Vision-language models trained on large, randomly collected data had significant impact in many areas since they appeared. But as they show great performance in various fields, such as image-text-retrieval,
their inner workings are still not fully understood. 
The current work analyses the true zero-shot capabilities of those models. We start from the analysis of the training corpus 
assessing to what extent (and which of) the test classes are really zero-shot and how this correlates with individual classes performance.
We follow up with the analysis of the attribute-based zero-shot learning capabilities of these models, evaluating how well this classical zero-shot notion emerges from large-scale webly supervision.
%
We leverage the recently released LAION-$400$M data corpus as well as the publicly available pretrained models of CLIP, OpenCLIP, and FLAVA, 
evaluating the attribute-based zero-shot capabilities on
CUB and AWA2 benchmarks. 
Our analysis shows that: 
%
(i) most of the classes in popular zero-shot benchmarks are observed (a lot) during pre-training; 
(ii) zero-shot performance mainly comes out of models' capability of recognizing class labels, whenever they are present in the text, and a significantly lower performing capability of attribute-based zero-shot learning is only observed when class labels are not used; (iii) the number of the attributes used can have a significant effect on performance, and can easily cause a significant performance decrease.
%


\end{abstract}

%% file: 1_introduction.tex
\tvspace
\section{Introduction}

\begin{figure}[t]
  \includegraphics[width=0.9\linewidth]{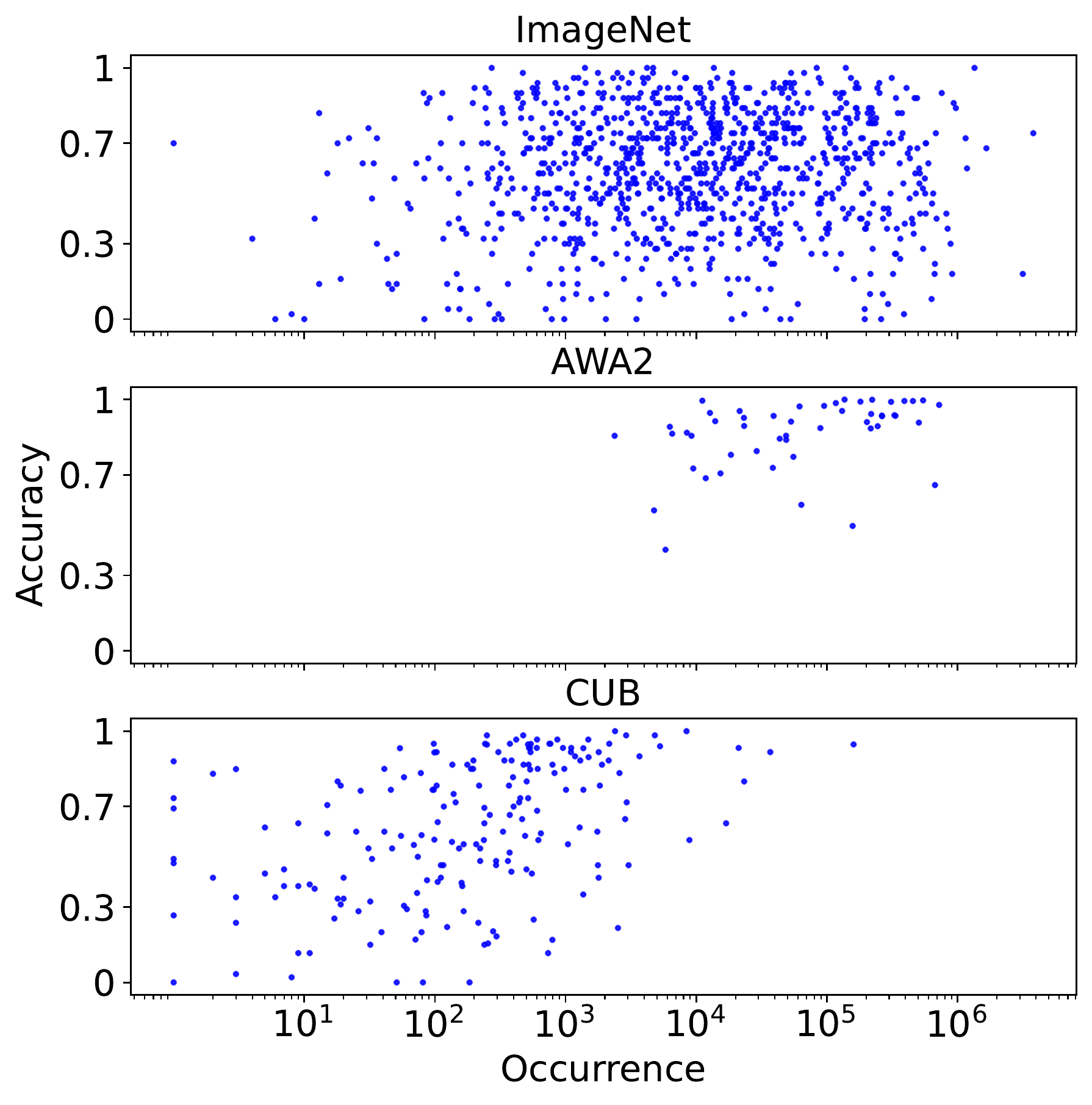}
  \nvspace
  \caption{Correlation between the occurrence of the ImageNet, CUB, AWA2 classes in the LAION-400M training samples (logarithmic x-axis) and the per-class recall of LAION-400M-pretrained OpenCLIP.}
  \label{fig:zero-shot-clip-Imagenet}
  \nvspace
\end{figure}

Models that are able to connect visual as well as linguistic information have been an active and growing research area for more than a decade~\cite{ferraro2015survey}.  
Based on the release of powerful architectures and models such as transformers~\cite{vaswani2017attention}, BERT~\cite{devlin2018bert}, T5~\cite{raffel2020exploring}, or GPT~\cite{radford2018improving} the last years, as well as with the availability of large-scale self-supervised learning methods, the area has seen a tremendous increase in performance.
Vision-language models pre-trained large-scale web-mined data, like CLIP~\cite{radford2021learning}, DALL-E~\cite{ramesh2021zero}, FLAVA~\cite{singh2022flava}, or ALIGN~\cite{jia2021scaling} and many more currently lead or at least underlay state-of-the-art methods in both discriminative and generative domains.  

One of the most interesting applications here is the case of zero-shot learning, which allows a model that has been trained on random web data to classify e.g. images by creating a text prompt that includes a meaningful class name such as ``A photo of a dog'' and classifies a set of images based on the distance of the embedding of the test text prompt and the embedding of the reference images~\cite{radford2021learning,ju2021prompting}. 
``Zero-shot'' in this case refers to the fact that the model has never been trained for this specific class or dataset, but does not account for the question of how often the class has been present in the training data. 
This setup deviates from the attribute-based zero-shot learning concept~\cite{xian2018zero}, which usually does not use any class information, but instead works on attribute descriptions such as ``brown, furry, 4-legged, strong'' to e.g. describe a bear. 
In this scenario, the target classes are not known and images can only be mapped to a certain class via their attribute descriptions. 
This paper addresses the question of how much zero-shot learning current web-scale trained vision language models are actually capable of and bridges the gap between the learning of semantic entities and attribute-based recognition. 

To motivate this problem, we leverage the recently released LAION-$400$M~\cite{schuhmann2021laion} data corpus, which is supposed to mimic the no-public WebImageText dataset~\cite{radford2021learning} that was used for the training of CLIP.  We count how often a class name actually appears in the LAION-$400$M dataset for three public datasets, namely ImageNet~\cite{russakovsky2015imagenet}, AWA2~\cite{xian2018zero}, and CUB~\cite{wah2011caltech} and evaluate the ``zero-shot'' classification performance of LAION-$400$M-pretrained OpenCLIP model~\cite{ilharco_gabriel_2021_5143773}.
As Figure~\ref{fig:zero-shot-clip-Imagenet} shows, the majority of Imagenet classes appears between 100 and 100K times in a $400$M samples web dataset.
While it can therefore not directly be compared to a training with a hand-curated dataset, it still leaves the assumption that common concepts like ``lion'' are not inferred from a high-level combination of vision-language cues, but rather directly learned from the data.
We further consider classes with extremely low as well as extremely high string occurrences for which the model performs significantly better or worse than average in Table~\ref{table:Clip_corner_cases}. 
Overall, we find that classes with high occurrences and low performance are semantically ambiguous, such a crane or nail.
Compared to that, for the case of low occurrence, high performance classes usually have a very distinct appearance as in the case of the jacamar and/or are actually special forms of more general classes like the mountain tent or the head cabbage. Another interesting observation is that they also can have descriptive names such as the ``black and gold garden spider`` which gives visual hints in the description.   

This motivates the question to which extent such models are able to navigate the class space compared to the attribute space of textual descriptions. To address this question, we leverage two common attribute-based datasets, the AWA2~\cite{xian2018zero} and CUB~\cite{wah2011caltech} and explore the effect of attribute information on current web-scale vision language data. 

The contributions of this paper are as follows: 
\begin{itemize}
\vspace{-2mm}
    \item We are the first to broadly analyze the zero-shot capabilities of web-trained vision language models with respect to class occurrence in the pre-training data; as well as the in-depth effects of the class and attributes combinations.
\vspace{-2mm}
    \item We propose VL-Taboo, a setup for the in-depth analysis of attribute learning capabilities of vision-language models.
\vspace{-2mm}
    \item We show that current VL models are suitable for zero-shot attribute-based classification, but that the overall performance is significantly lower than a label-based classification. 
\vspace{-2mm}
    \item We show that, given the class label, attributes have significantly less impact, but that additional descriptions, which also lead to longer prompts, can have an effect on some models.
\end{itemize}
\vspace{-2mm}

We will make the code base of the proposed project publicly available in form of an attribute-VL gym to allow for an easy adaptation to any vision-language models (\href{https://github.com/felixVogel02/VL-Taboo/tree/main}{VL-Taboo}).










\begin{table}[t]
\footnotesize

\begin{tabular}{ |p{5.4cm}|p{0.6cm}|p{0.6cm}|}
\hline
\multicolumn{3}{|c|}{Low occurrence ($<100$), high recall ($>0.75$)} \\
\hline
Class labels& occ. &recall\\
\hline
black and gold garden spider, Argiope aurantia &82 & 0.9\\
jacamar & 87& 0.86 \\
European gallinule, Porphyrio porphyrio &18 &0.7\\
mountain tent &91 & 0.88 \\
head cabbage & 31 & 0.76  \\
\hline
\end{tabular}

\begin{tabular}{ |p{5.4cm}|p{0.6cm}|p{0.6cm}|  }
\hline
\multicolumn{3}{|c|}{High occurrence ($>10^5$), low recall ($<0.05$)} \\
\hline
Class labels& occ. &recall\\
\hline
crane &103K&0.0\\
nail & 389K &0.02 \\
tub, vat &195K &0.0 \\
ear, spike, capitulum & 260K  &0.0  \\
\hline
\end{tabular}


\caption{Extreme values of occurrence and recall from OpenCLIP trained on LAION-400M evaluated on ImageNet.}
\nvspace
\label{table:Clip_corner_cases}
\end{table}


%% file: 2_relatedwork.tex
\tvspace
\section{Related work}
\subsection{Web-scale Vision Language Models}

Within the last years an abundance of vision-language architectures has been proposed (Table~\ref{table:overviewVLmodels}). For a complete overview, we refer to Chen et al.~\cite{Chen2022VLPSurvey} and Long et al.~\cite{Long2022ijcaisurvey}. Especially in the recent two years, more and more models emerged that are trained on web data only. Such training datasets are more noisy, compared to human-generated image-text pairs, but also easy to acquire and scale. Such models can therefore be trained on e.g. 70M image-text pairs as in the case of FLAVA~\cite{singh2022flava} up to more than one Billion in the case of ALIGN~\cite{jia2021scaling} and SimVLM~\cite{wang2021simvlm}. The problem regarding a deeper investigation of those models is that often, neither the model itself nor the training data is publicly available. 
To address this fact, Wortsman et al. recently published the Open-CLIP model~\cite{ilharco_gabriel_2021_5143773} together with the Laion-$400$M dataset~\cite{schuhmann2021laion}, that allows for an in-depth analysis of those architectures, also with respect to the training data. To allow for the here presented study we therefore rely on the appearance of training samples in the Laion-$400$M dataset as a reference distribution for large-scale web-mined data. In terms of publicly available backbones that are trained on web-scale data, we further identified the well know CLIP~\cite{radford2021learning} model, as well as the FLAVA~\cite{singh2022flava} model. Note that we also tested on the model with the next bigger training set VILT~\cite{kim2021vilt} with 10M samples, but found that it has low performance for class-only prompting, such as the 4.4 recall on CUB~\cite{wah2011caltech} dataset, which indicates that the training set size and source, even if it is noisy data, does play a significant role for the zero-shot capabilities of such models.  






\begin{table}[t]
\centering
\resizebox{0.9\columnwidth}{!}{

\begin{tabular}{p{1.8cm} p{0.8cm}| p{2cm} p{0.8cm} p{0.8cm} }
\toprule
\multicolumn{2}{c|}{Model} & \multicolumn{3}{c}{Dataset} \\
\hline
Name&Public & Name& Size &Public \\
\hline
OpenCLIP~\cite{wortsman2022robust} & \cmark & Laion-$400$M & 400M & \cmark  \\
CLIP~\cite{radford2021learning} & \cmark & WebImageText & 400M & \xmark \\
FLAVA~\cite{singh2022flava} & \cmark & PMD & 70M & \cmark   \\
\hline
ALIGN~\cite{jia2021scaling} & \xmark & JFT & 1.8B & \xmark \\
SimVLM~\cite{wang2021simvlm} & \xmark & JFT & 1.8B & \xmark  \\
Florence~\cite{Yuan2021Florence} & \xmark & FLD-900M & 900M & \xmark  \\
FILIP~\cite{yao2022filip} & \xmark & FILIP300M & 300M & \xmark  \\
\hline
ViLT~\cite{kim2021vilt} & \cmark & Combination & 10M & \cmark \\
VinVL~\cite{zhang2021vinvl} & \cmark & Combination & 9M & \cmark   \\
Oscar~\cite{li2020oscar} &  \cmark& Combination & 5.65M &\cmark\\
ALBEF~\cite{li2019visualbert} & \cmark & Combination & 5M & \cmark \\
TCL~\cite{yang2022vision} &\cmark  & Combination & 5M & \cmark\\
UNITER~\cite{chen2019uniter} &  \cmark &  Combination& 4.7M & \cmark \\
LXMERT~\cite{tan2019lxmert} & \cmark & Combination & 920k &\cmark\\
\bottomrule
\end{tabular}
}
\nvspace
\caption{Overview of vision-language models and their availability with respect to checkpoints and training data. 
\label{table:overviewVLmodels} }
\nvspace
\end{table}

\tvspace
\subsection{Evaluation of VL models}

With the plethora of available vision language models, research also becomes more and more interested in comparative evaluations of those models. 
In this context, VLUE benchmark~\cite{zhou2022vlue} by Zhou et al. was proposed to evaluate the
generalization capabilities of VL models by testing them on a diverse dataset collected across different cultures and countries and annotated by native speakers. 
~
The VALSE benchmark~\cite{parcalabescu2021valse} by Parcalabescu et al. was proposed to compare the capabilities of the different VL models to learn specific linguistic phenomena, such as the ability to count objects, understand the presence of objects on the image, understand objects' relations or actions, etc. Similarly, VL-CheckList benchmark~\cite{zhao2022vl} by Zhao et al. tests the abilities of VL models to understand objects, their attributes and relations. Other works e.g. by Hendricks et al.~\cite{hendricks2021probing} evaluate the capabilities of VL models to understand verbs.
Some benchmarks~\cite{kayser2021vil,bitton2022winogavil} were also proposed to assess the reasoning skills of VL models, such as the e-VIL benchmark~\cite{kayser2021vil} which focuses on analyzing the capabilities of VL models to generate natural language explanations for their predictions, which requires an understanding of complex concepts and performing higher-order reasoning.
In this context, the WinoGAViL benchmark~\cite{bitton2022winogavil} was proposed to challenge commonsense reasoning and association abilities of VL models, e.g. associating the word "werewolf" with an image of ``dog'' or an image of ``full moon''.

Our benchmark complements existing frameworks by analyzing of abilities of VL models to employ information in language description with respect to utilizing an object label and multiple additional object attributes.  Compared to VALSE~\cite{parcalabescu2021valse} and VL-checklist~\cite{zhao2022vl} benchmarks, which tests the abilities of VL models to understand object and attributes, considering language descriptions with correct or wrong object label or attributes, we focus on how much models utilize object label and attributes while computing text embeddings, and how much it influences classification performance. This also includes the question if the long language description with many attributes helps to recognize the correct image, and how much for model relies on the object label over that on given attributes.

\tvspace
\subsection{Attribute based (zero-shot) learning}
Compared to the recent success of vision language models, attribute-based zero-shot learning itself can be considered a much older research topic, e.g. been formulated in 2008 by Larochelle et al.~\cite{larochelle2008zero} as the problem of recognizing classes without available training where only descriptions of these classes/tasks are given. The paper also addresses the aspect of attribute-based descriptions and defines
zero-shot learning as the abbility of inferring new classes at test time, which the model has not seen before e.g. based on a given set of attributes per class or per image. 
Based on that follow-up works, e.g. by Lampert et al.~\cite{lampert2013attribute} proposed different architectures for attribute-based recognition, as well as respective attribute-based datasets such as "Animals with Attributes" and "Animals with Attributes 2" \cite{xian2018zero} consisting of 50 classes where the first 40 classes are used for training and the last 10 classes for testing. Akata el. al.~\cite{akata2015label} further utilized CUB-200-2011 (CUB)~\cite{wah2011caltech} dataset with 200 different bird species for attribute-based recognition with 150 training classes and 50 validation classes.
Note that those benchmarks are mainly created for models trained or fine-tuned on an attribute-based dataset and not for vision-language models. Our results can therefore not directly be compared to state-of-the-art zero-shot systems such as e.g. TransZeros by Chen et al.~\cite{chen2022transzero} or MSDN~\cite{Chen_2022_CVPR}.


%% file: 3_method.tex
\tvspace
\section{System description}

In the following, we give a short overview of the design of the evaluated architectures and the evaluation setup.

\subsection{Models}

\paragraph{CLIP / OpenCLIP}
First we consider both, the CLIP~\cite{radford2021learning} as well as the OpenCLIP model~\cite{ilharco_gabriel_2021_5143773}. As OpenCLIP follows CLIP with respect to architecture and training, we discuss both together.
In both cases, the image encoder and text encoder are trained from scratch.
~
The text encoder is based on a transformer architecture
that incorporates the modifications described in~\cite{radford2019language}
and operates on a vocab size of 49,152 \cite{sennrich2015neural}.
The maximum input sequence length is limited to 76.
~
For the image encoder, five ResNets and three Vision Transformer models are provided. 
All models are trained on 32 epochs on the WIT dataset~\cite{radford2021learning} (400M image-text pairs) with a batch size of $N=32,768$, maximizing the cosine similarity between the $N$ correct image-text pairs (of the $N \times N$ possible pairings in a batch) and minimize the cosine similarity between the incorrect $N^2 - N$ pairings with Noise Contrastive Estimation~\cite{oord2018representation}. 
Based on that it learns a multi-modal embedding space by jointly training an image and text encoder via contrastive learning. 
For our analysis, we use ViT-B/32 as the image encoder that generates vector representations of images of size 512, as does the text transformer.

\tvspace
\paragraph{FLAVA}
Compared to CLIP and OpenCLIP, FLAVA~\cite{singh2022flava} is optimized on multiple vision, language, and cross- and multi-modal tasks. 
The FLAVA model contains an image encoder and a text encoder to extract unimodal image and resp. text representations, followed by a multimodal encoder that fuses and aligns the image and text representations for higher-level tasks. 
Both image and text encoder follows a ViT-B/16 architecture that outputs a list of state vectors. 
For the multimodal encoding, the two lists of state vectors are further forwarded to a transformer module that is based on the ViT architecture.
Different from CLIP and OpenCLIP, FLAVA is trained with respect to several multimodal and unimodal objectives, including global contrastive loss, that similarly to CLIP contrasts unimodal image and text representations, masked multimodal modeling, image-text matching, and masked image and mask language modeling.
We use the unimodal image and text encoders that are trained with contrastive loss and compute image-text similarities in multi-modal 768-dimensional embedding space. 

\subsection{Evaluation setup}

\paragraph{Classification} During evaluation, for each image, we create \textit{c} prompts, where \textit{c} is the number of classes in dataset~\textit{d}, resulting in one textual prompt for each class.
The prompts are tokenized and encoded by the model \textit{m} and the resulting vector is L2-normalized. The image $i\in d$ is also encoded by \textit{m} and L2-normalized. We then calculate the cosine similarities between the processed image \textit{i} and the \textit{c} prompts and apply softmax to the similarity values. The class that corresponds to the prompt that is most similar to image \textit{i} is the predicted class. The predicted class and the label class are compared for each image and the average accuracy over \textit{d} is reported. 

\tvspace
\paragraph{Prompting}
For a prompt generation, all prompts start with: \textit{a photo of a $<$class label$>$}, or when the class label is not used: \textit{a photo of an animal} for the AWA2 dataset~\cite{xian2018zero} or \textit{a photo of a bird} for CUB dataset~\cite{wah2011caltech} that contains images of different bird species.
We concatenate the initial prompt with the respective image attributes. When we add an attribute with attribute description \textit{attrDesc} and expression \textit{expr} to the prompt, creating the new prompt as follows: \textit{$<$initial prompt$>$ + that + $<$attrDesc$>$ + $<$expr$>$}. Attribute descriptions and expressions depend on the dataset.

%% file: 4_datasets.tex
\tvspace
\section{Datasets}


\paragraph{CUB}
The Caltech-UCSD Birds-200-2011~\cite{wah2011caltech} dataset contains 11,788 images of 200 different bird species.
Each image is annotated by 312 binary attributes with certainty annotations.
For the following evaluation, we combine the training and test set and always evaluate on 11,788 images.

Attributes that are provided by CUB for each image separately and consist of an attribute description and an expression. For example, one attribute can have the description \textit{attrDesc}: \textit{``has head pattern''} with the expression \textit{expr}: \textit{``crested''}. Each attribute description has different possible expressions. 
CUB assigns attributes with certainty levels. 
We only use attributes that are assigned to the image with a certainty level of \textit{probably} or \textit{definitely} skipping attributes with lower certainty levels.

CUB also provides class attributes. For each class and attribute, a probability is given that the attribute can be found in an image of that class.  For each attribute description, we take a maximum of one attribute with the highest probability if its probability is also over 0.


\tvspace
\paragraph{AWA2}
Animals with Attributes 2 (AWA2)~\cite{xian2018zero} consists of 37,322 images of 50 animal classes. 85 binary attributes are annotated for each class, differently from CUB that has per image attribute annotation. 
Note that we again use the full dataset, train, and test classes, for evaluation.
In contrast to CUB, the attributes of AWA2 do not have separated attribute descriptions and expressions, but mainly feature one or two-word attributes which can be verbs, adjectives, adverbs, or nouns. 
We therefore add the attributes as a comma-separated list at the end of the prompt.
In order to stay comparable to CUB, each prompt starts with: \textit{a photo of an animal}, or \textit{a photo of a $<$class label$>$} and we add \textit{attrDesc}: \textit{``with attribute''} at the end of the prompt. Every other attribute gets added with a comma.



\tvspace
\paragraph{Attribute detection for AWA2}
Compared to CUB, AWA2 provides only class attributes not image attributes. 
For example for the class \textit{rabbit} different colors are assigned: \textit{black}, \textit{white}, \textit{brown}, and \textit{gray}, although a rabbit typically has only one or two colors. 
As the analysis is based on attributes that are present per image
we need to find a subset of the class attributes that fit an image. 
We do this for each model $m\in \{OpenCLIP,~CLIP,~FLAVA\}$ separately. 
For each attribute \textit{z} of $c$ attributes assigned to the class we create three prompts:
 \textit{a photo of a $<$z$>$ animal}, \textit{a picture of a $<$z$>$ animal}, and \textit{a photograph of a $<$z$>$ animal}.
We encode each prompt with the model $m$ and take the average of the three encoded prompts and L2-normalize it. 
The cosine similarity between the image and the $c$ vectors representing the $c$ attributes of the class is calculated.
We keep all attributes that have a similarity to the image that is higher than a cut-off value which is chosen individually for each model $m$, preserving on average $20$ attributes per class.

%% file: 5_experiments.tex
\tvspace
\section{VL-Taboo}

\label{sec:class_label_eval}
\begin{figure}[t]
  \includegraphics[width=1\linewidth]{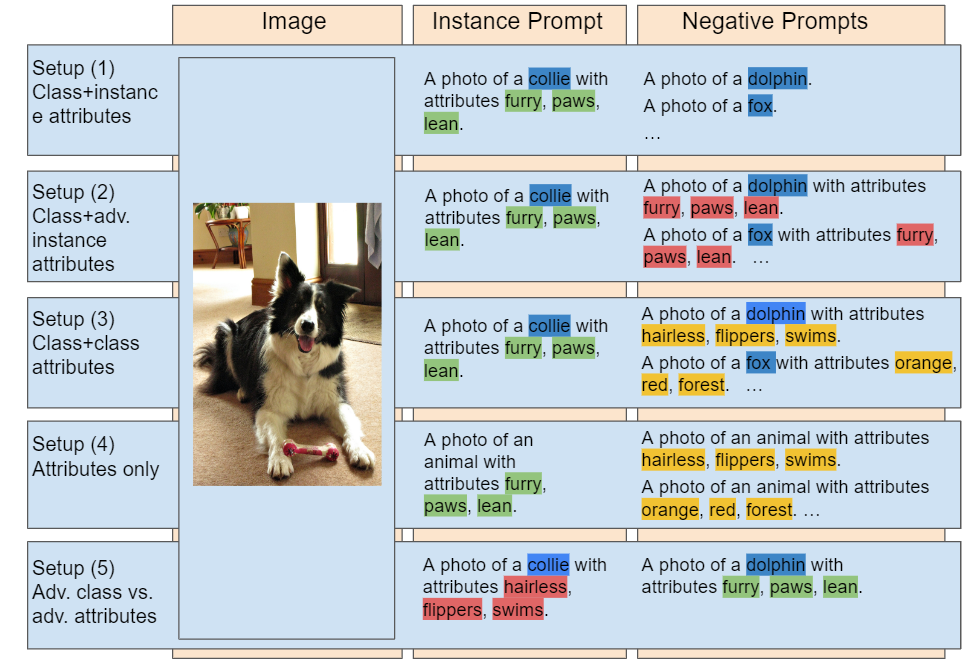}
  \nvspace
  \caption{\textbf{Example prompts for Setups (1)--(5)} with three attributes on the AWA2 dataset. The instance prompt is created the same way in \textit{Setups (1)--(3)}: it contains the class label and image attributes. The prompts for other (negative) classes are created differently. In \textit{Setup~(1) Class + instance attributes}, only the class label is used. In \textit{Setup~(2) Class + adversarial instance attributes}, the class label and the same image attributes are used. While in \textit{Setup (3) Class + class attributes}, the class label and class attributes are used. In \textit{Setup~(4) Attributes only}, the instance prompt is created without the class label and only by image attributes. Prompts for negative classes are created with only class attributes and no class label. In \textit{Setup~(5) Adversarial class vs. adversarial attributes}, only two prompts are compared. One prompt is with the correct class label and wrong attributes, another prompt is with an incorrect class label but correct image attributes. Class labels are highlighted \textit{blue}, instance attributes \textit{green}, attributes that don't belong to the class \textit{red} and class attributes \textit{yellow}.}
  \nvspace
  \label{fig:sample_sentences}
\end{figure}




In the following, we describe our VL-Taboo benchmark for a detailed attribute-based analysis of VL models. In the VL-Taboo benchmark, we investigate the influence of textual attributes on the accuracy of image-text retrieval under two different conditions. 
First, in Section~\ref{sec:class_label}, we consider the case where the class label is given at the beginning of each prompt, and attributes are added to provide a more detailed description of the object. 
Second, in Section~\ref{sec:no_class_label}, we exclude the class labels from the prompts and only the attributes-based description is given. 
We consider five setups overall and discuss them in detail in the following. 
An example of all setups is given in Figure~\ref{fig:sample_sentences}. 
For each image, we test for different numbers of attributes, assuming that the description gets more detailed whit each attribute added.


\tvspace
\subsection{Classification Performance with Class Label}
\label{sec:class_label}
\begin{figure*}[t]
   \centering
   \includegraphics[width=0.75\linewidth]{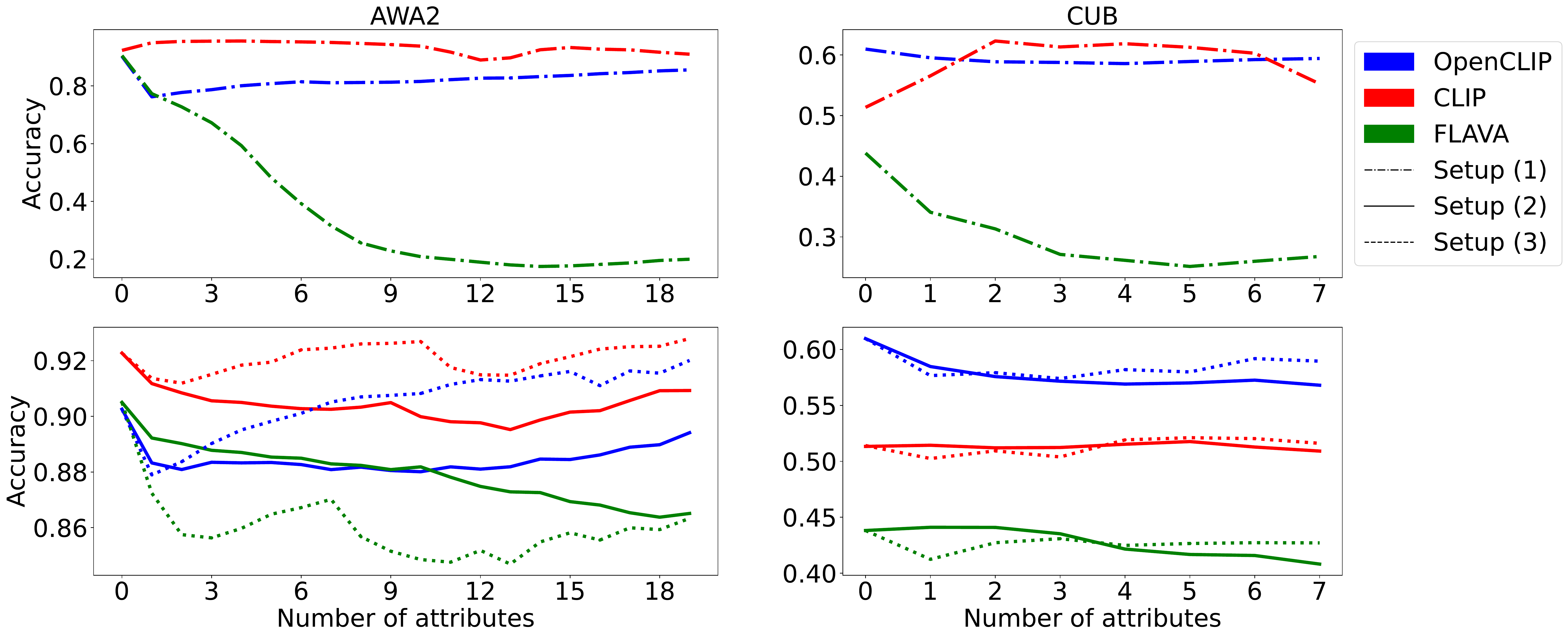}
%
  \nvspace
  \caption{\textbf{Evaluation CLIP, OpenCLIP, FLAVA models on AWA2 and CUB datasets in Setups (1)--(3).} Average accuracy when attributes are added to the class label. We consider three cases: in \textbf{Setup (1): Class + Instance attributes}, attributes are only added to the instance prompt; in \textbf{Setup (2): Class + adversarial attributes}, the same image attributes are added to all class prompts as adversarial description resulting in more similar prompts with each attribute added; and in \textbf{Setup (3): Class + Class attributes}, image attributes are added to the instance prompt and correct class attributes are added to each class label, making all prompts more distinctive. Best viewed in color and zoom. \label{fig:exp123_awa2_cub}}
  \nvspace
\end{figure*}

\begin{figure*}[t]
  \centering
  \includegraphics[width=1\linewidth]{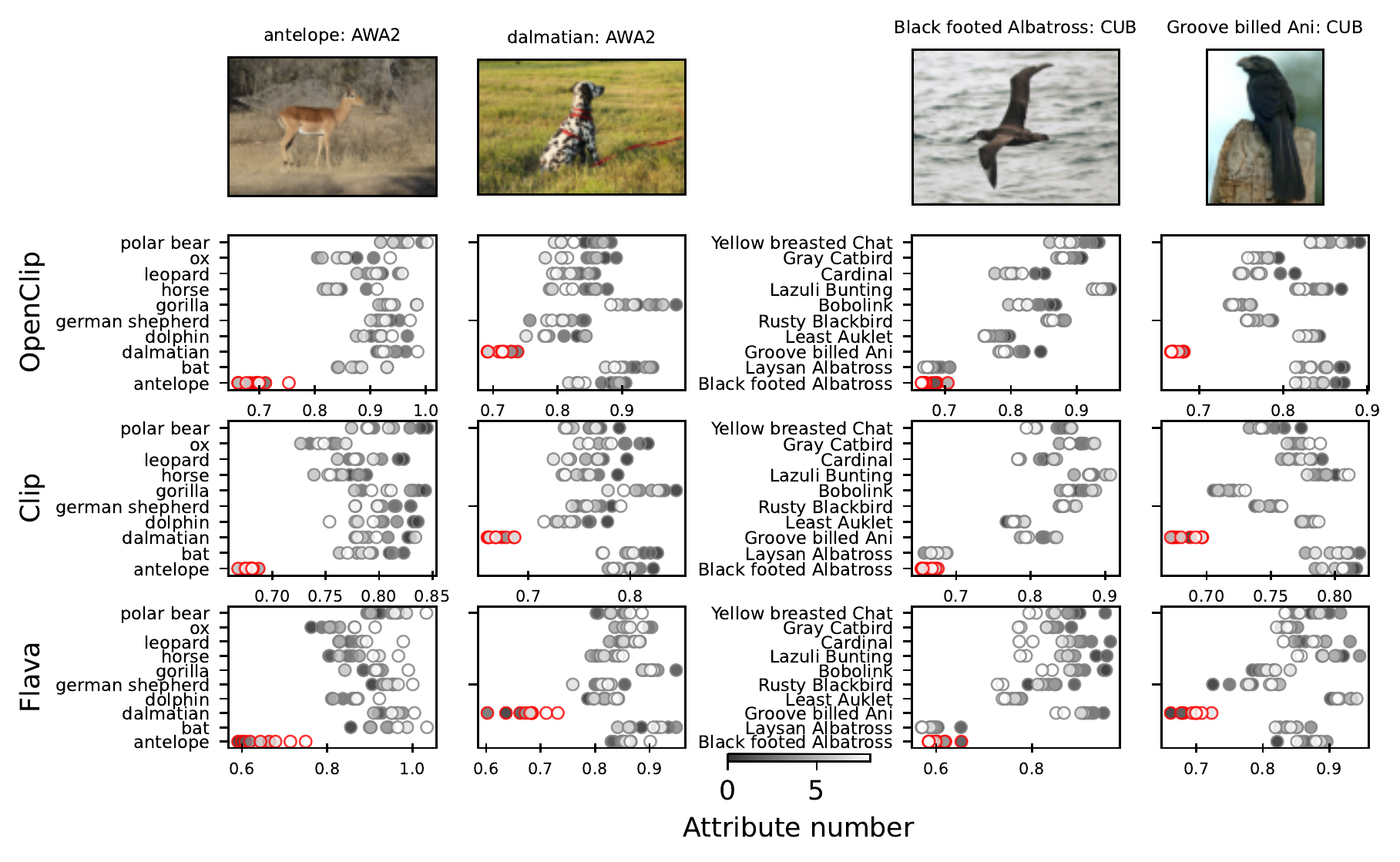}
  \nvspace
  \caption{\textbf{Distance of image encoding and query prompts for OpenCLIP, CLIP, and FLAVA on CUB and AWA2.} Each scatter plot shows distances of an image encoding (top of each column) and a set of prompts. Each row in a scatter plot consists of 9 dots that correspond to a similarity with 9 different prompts. Each prompt is constructed from a class name (y-axis label) and (according to Setup \#2) 0-8 ground truth attributes for the reference image instance (top of each column). The more attributes a prompt contains, the lighter the dot is colored. X-axis corresponds to the cosine distance of the prompts to the image encoding. The prompts that contain the true class label of the reference image are colored red. \label{fig:cosdist_exp2}}
  \nvspace
\end{figure*}

\begin{figure*}[t]
\centering 
    \begin{subfigure}{0.7\linewidth}
        \begin{subfigure}{0.48\linewidth}
            \includegraphics[width=1\linewidth]{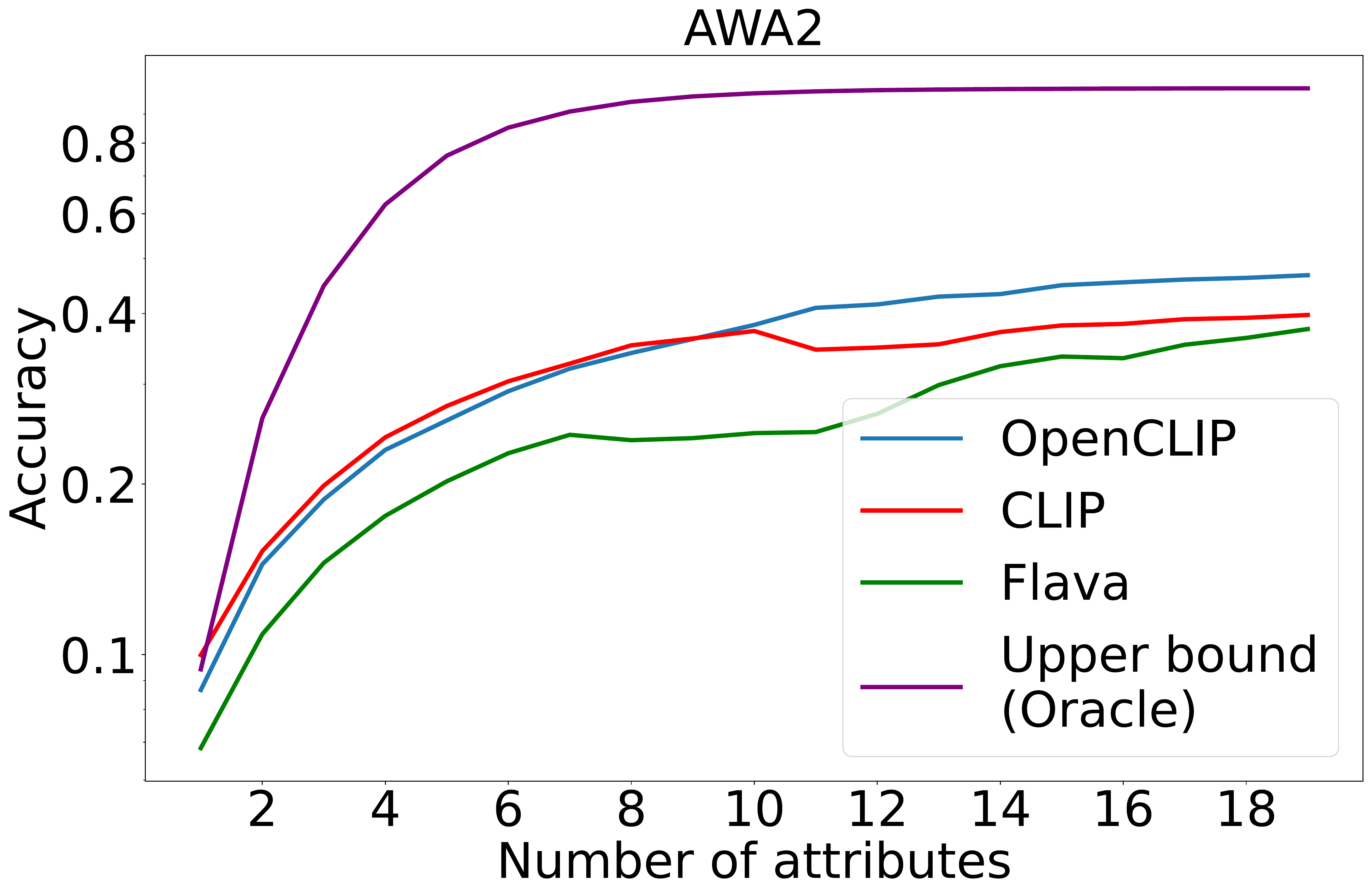}
        \end{subfigure}
        \begin{subfigure}{0.48\linewidth}
            \includegraphics[width=1\linewidth]{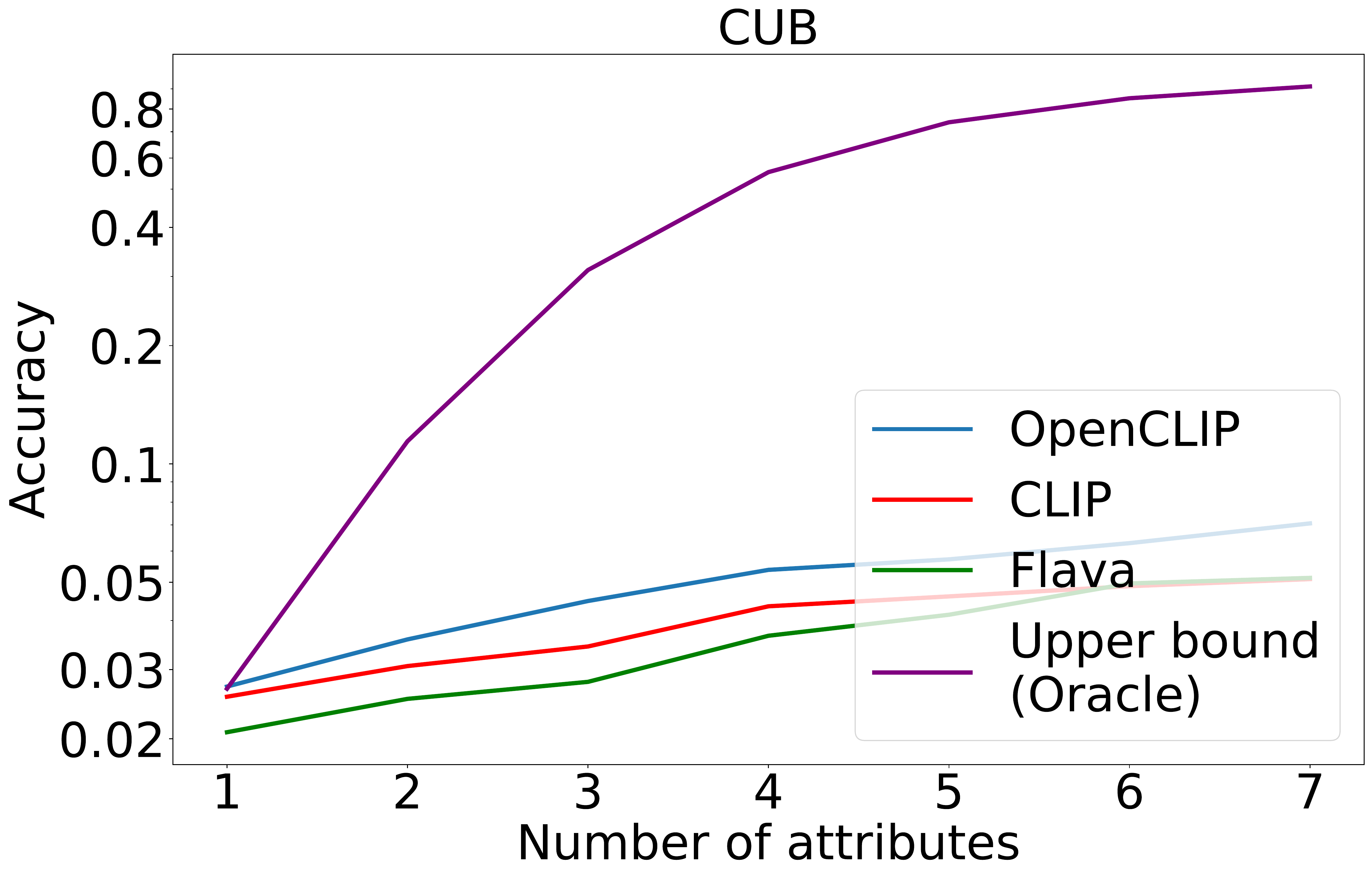}
        \end{subfigure}
    \end{subfigure}

%
  \nvspace
  \caption{\textbf{Evaluation CLIP, OpenCLIP, FLAVA models on AWA2 and CUB datasets in Setup (4): Attributes only.} 
  Average accuracy for classification with the prompts generated with attributes, but \textit{without the class label.} 
  For an image, we compare one prompt that contains image attributes with prompts for every other class with class attributes. \label{fig:exp4_awa2_cub}}
  \nvspace
\end{figure*}

\begin{figure*}[t]
     \centering 
    \begin{subfigure}{0.7\linewidth}
        \begin{subfigure}{0.48\linewidth}
            \includegraphics[width=1\linewidth]{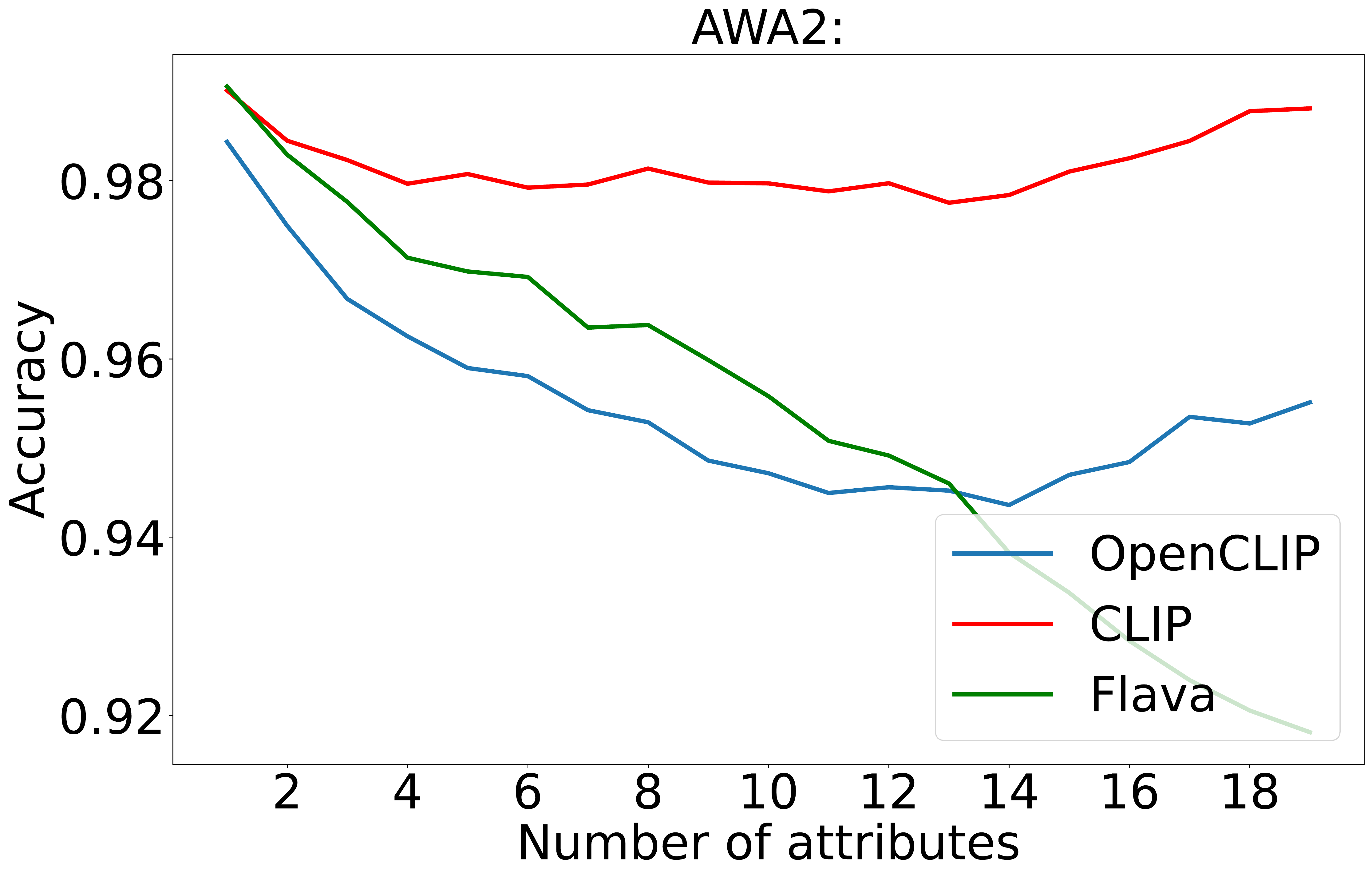}
        \end{subfigure}
        \begin{subfigure}{0.48\linewidth}
            \includegraphics[width=1\linewidth]{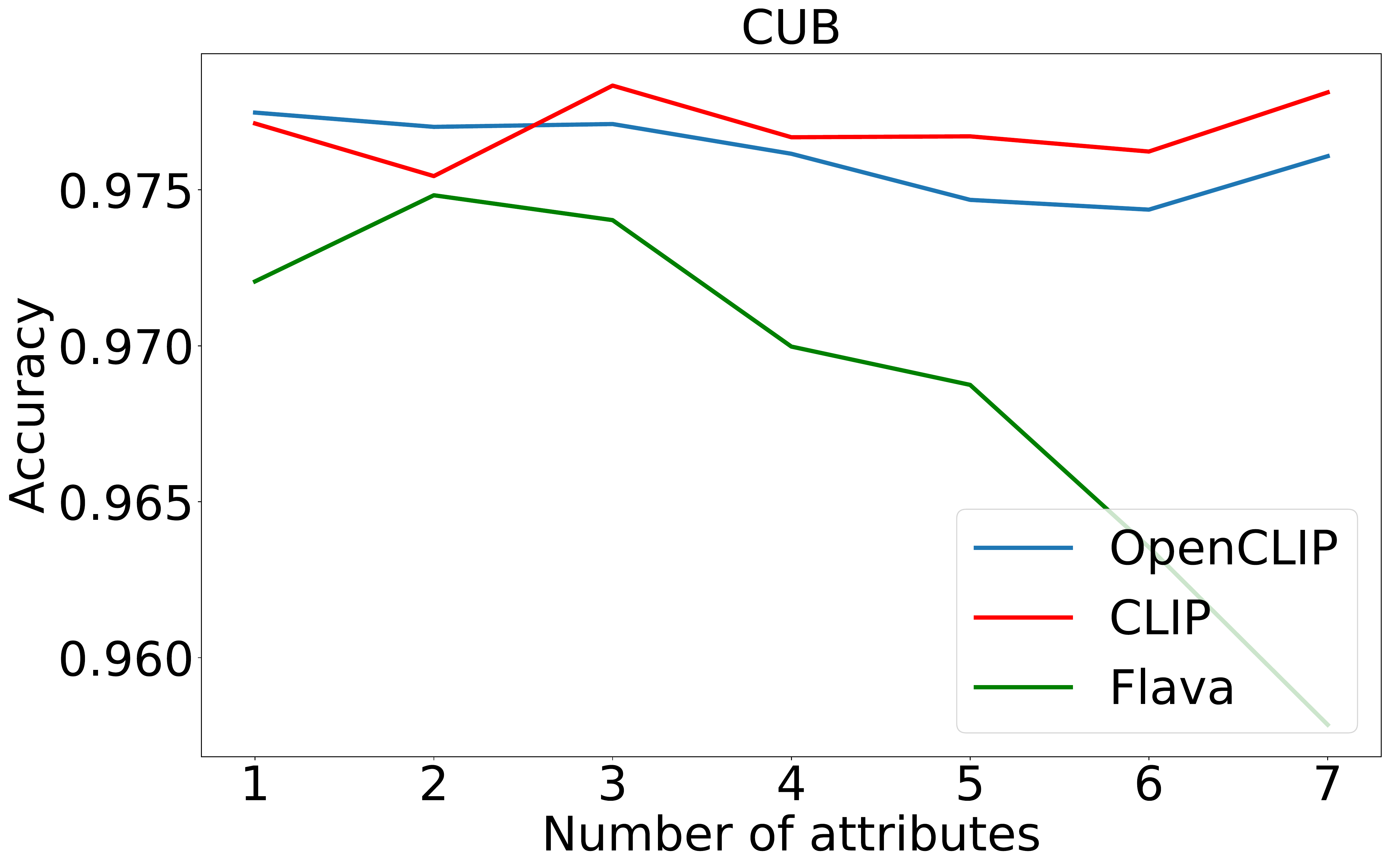}
        \end{subfigure}
    \end{subfigure}
%
  \nvspace
  \caption{\textbf{Evaluation CLIP, OpenCLIP, FLAVA models on AWA2 and CUB datasets in Setup (5): Adversarial Attributes vs Adversarial Class.} 
  For each image two prompts are created. The first prompt contains the correct class label of the image and only wrong attributes. The second prompt contains a wrong class label and only correct image attributes. The average percentage of choosing the correct label with the wrong attributes over the wrong label with the correct attributes. \label{fig:setup5_awa2_cub}}
  \nvspace
\end{figure*}


\paragraph{Setup (1): Class + Instance attributes}
In the first setup, we consider the case that the prompt that includes the correct class label and additional image attributes is compared to prompts with other class labels (Figure~\ref{fig:sample_sentences}), thus the instance prompt becomes more descriptive, while the negative class prompts stay the same. 
The results are shown in Figure~\ref{fig:exp123_awa2_cub} (dashed line). 
Note that the results for zero attributes also present the baseline for class-only prompting. 
It shows that OpenCLIP actually drops slightly and that FLAVA drops significantly when more attributes are added.  
Only CLIP profits from the additional information provided via the attributes, but this observation only holds for a limited number of attributes. As the number of attributes grows, also CLIP results tend to drop below the baseline performance. Overall, it can be concluded that \textit{the length} of the instance prompt, which increases with every attribute added, has a more significant impact than \textit{the additional signal} provided by a more detailed description, especially in the case of FLAVA. To verify this, we plot the absolute distance values of the prompts to the respective image embedding in Figure~\ref{fig:cosdist_exp2}.  
Looking only at the distance of the reference instance (red), we can observe that adding more attributes influences the instance embedding most for FLAVA, while the embedding of CLIP and OpenCLIP stays more compact. Additionally, we see that in three out of four cases, adding more attributes results in prompt embeddings increasing distance from the original image embedding, which can not be observed for the two other models.





\tvspace
\paragraph{Setup (2): Class + adversarial attributes }

In the second setup, we compare the instance prompt with the class label and the correct image attributes with prompts with other class labels, but keep the image attributes fixed. 
Practically, this means that, as the prompts get longer, they also get more similar as they only differ with respect to the class label.  
Thus, compared to setup (1), the prompt that belongs to the image class stays the same, but all the other prompts contain the same attributes, while in setup (1) the other prompts do not contain any attributes.
As a result, all prompts that are compared have the same length. 

Looking at the results in Figure~\ref{fig:exp123_awa2_cub} (solid line), it shows that the added words help to stabilize the performance in all three models and that there are now only slight deviations from the initial baseline performance. 
Namely, the performance for CLIP does no longer increase, but also not significantly decrease, and in the case of FLAVA even completely removes the performance drop of the first setup. Looking at the embedding distances for this setup in Figure~\ref{fig:cosdist_exp2}, we see that in the case of CLIP and Open CLIP, the embeddings with more attributes (lighter dots) move closer to the reference image embedding, but that there is still a margin between the farthest away instance embedding (red) to all other negative embeddings. It can therefore be assumed, that the additional attributes are recognized in the image, but that the class label provides a stronger signal compared to the significantly larger amount of attributes. 
For the case of FLAVA, we see that the attributes also change the embedding, but they sometimes decrease and sometimes increase the distance from the reference image. What seems important here is that the tendency of increasing or decreasing distance is consistent with the tendency of the instance prompt and therefore results in a similar performance, no matter if attributes are added or not. 





\begin{figure*}[t]
   \centering
  \includegraphics[width=0.98\linewidth]{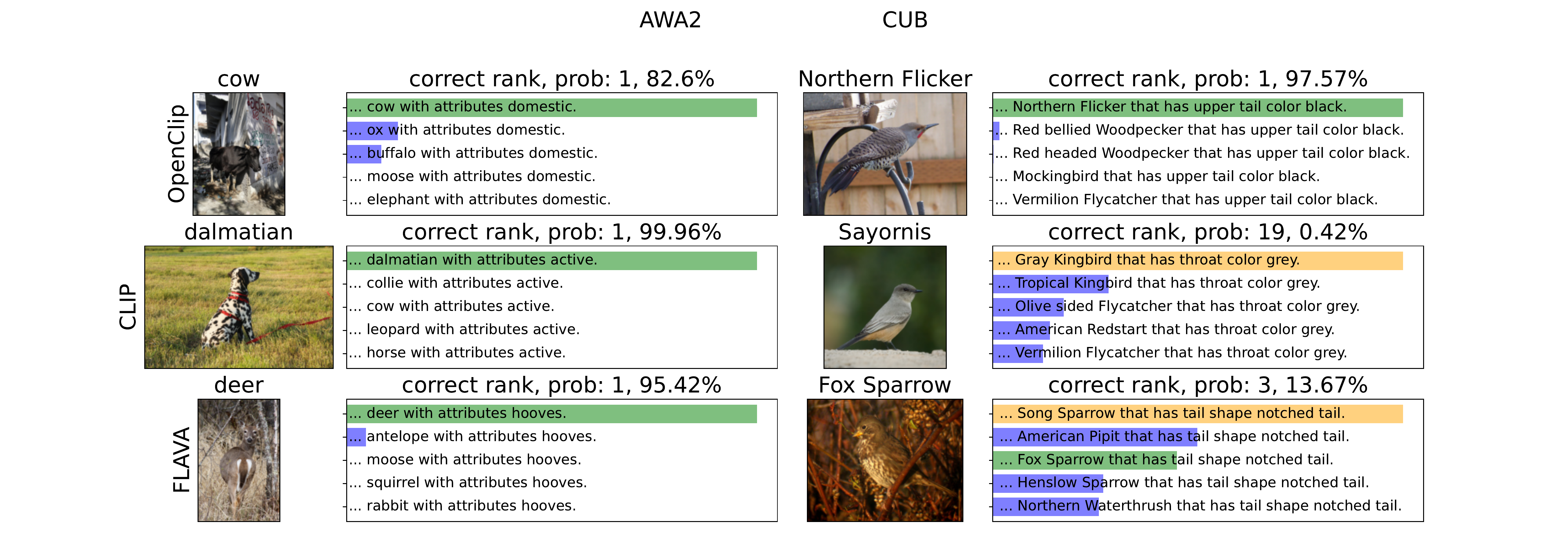}
  \nvspace
  \caption{\textbf{Qualtitative examples of OpenCLIP, CLIP, and FLAVA models on CUB and AWA2 datasets.} 
  Retrieving a text corresponding to an image encoding using different models. The text gallery set is formed by combining a class label (out of all possible classes) with a single ground truth attribute for each query image (Setup \#2). As can be seen in many cases texts corresponding to semantically similar classes are retrieved in the top options.
  \label{fig:qual_exp2}}
  \nvspace
\end{figure*}

\tvspace
\paragraph{Setup (3): Class + Class attributes}

In this setup, we assume the best-case scenario and compare the instance prompt with prompts generated with class labels and the respective correct class attributes.
As in setup (2), prompts have similar length, but attributes differ for each prompt, so only the instance prompt attributes fit the respective instance image, while all other attributes fit the respective other classes. 
The results in Figure~\ref{fig:exp123_awa2_cub} (dotted line) show that this setup is slightly beneficial for CLIP and OpenCLIP on AWA2 and also for CLIP on CUB, in which cases the systems are able to slightly increase performance compared to the baseline, but also that this performance increase is not strong. Looking again at the absolute distances depicted in Figure~\ref{fig:cosdist_exp2}, it can, similar to setup (2), be concluded, that mainly the class label defines the overall position in the embedding space so that adding more attributes, no matter if correct or not, does not necessarily change the relative ordering of distances in this case.  
That correct attributes do not necessarily change the embedding in a positive way can also be seen when directly comparing the results with adversarial attributes (dotted line) to the ones with correct class attributes (dashed line). Here, in the case of CUB, the correct labels are in the beginning outperformed by the adversarial ones, and it seems that the impact of correct attributes only becomes clear when four or more attributes are present.





\tvspace
\subsection{Attribute-based Classification Performance }
\label{sec:no_class_label}

\paragraph{Setup (4): Attributes only} While in the previous experiments, the class label was present in each prompt and the attributes only gave additional information, we now asses the traditional zero-shot case with only attributes.
For each image, we create one prompt with \textit{x} image attributes and without the class label. For each of the other \textit{c-1} classes, we choose class attributes randomly and independently. 
Note that this setup works exactly the same as setup (3) with the only difference that the class labels are left out. 
%
The results for attribute-only classification are shown in Figure~\ref{fig:exp4_awa2_cub}. It shows that the overall performance drops significantly compared to the results in case a class label is given. 
Since we only use attributes and no unique class labels it is possible that prompts that belong to other classes than the image class only include attributes that also show in the image. 
We therefore include the oracle upper bound that assumes ground truth knowledge of the image attributes to Figure~\ref{fig:exp4_awa2_cub}. 
%
Another important difference for all tested models is that adding more attributes actually improves performance. 
Even assuming different base chance ratios for both datasets, the improvement can be considered more significant for AWA2 compared to CUB, which might be based on the fact that the AWA2 animal classes as well as their attribute descriptions are more common compared to the expert knowledge in CUB and that the query prompts might more resemble the training data or relate to existing patterns in the data.

\tvspace
\paragraph{Setup (5): Adversarial Attributes vs Adversarial Class} %
To finally directly validate the importance of class labels compared to the importance of attributes we consider a binary adversarial setup. In this case, we create two prompts for each image, in which the first prompt contains the class label of the image but only attributes that are not present in the image and the second prompt contains a randomly chosen wrong class label, but only attributes that fit the image. Both sentences include the same number of attributes. 
We calculate the similarity between the image and both prompts and measure how many images the first prompt with the correct class label is predicted. The results are shown in Figure~\ref{fig:setup5_awa2_cub}. It shows that all models prefer the correct class with wrong attributes over the wrong class with correct attributes, but also that this effect is more pronounced for CLIP and OpenCLIP, while FLAVA shows a slight tendency to actually consider the correct attribute description the more attributes are added to the prompt. 

\tvspace
\subsection{Qualitative results}

To further support the results of the quantitative evaluation, we also consider qualitative examples from all three models for AWA2 and CUB, showing a sample image together with the respective top-5 retrieval results in Figure~\ref{fig:qual_exp2}. More qualitative results can be found in the supplementary. It shows that for AWA2, all models show a high probability for the correct class label, even under adversarial conditions but also relevant, related classes for the second and third option, whereas the results for CUB are less clear. 




%% file: 6_conclusion.tex
\tvspace
\section{Conclusion} 

This work presentes an in depth study of the attribute-based zero-shot capabilities of web-scale trained vision language models. To this end, we leveraged attribute annotated datasets and used them to create two settings to benchmark current models, one in which the class label is given and attributes are added as additional information and one where the effect of only attribute description is assessed. It shows that even while the class label itself makes only for a small fraction of the overall prompt, it can be considered the guiding information. It further shows that in a true attribute-based zero-shot learning, the models are still able to classify the data, but also that the main information is drawn from the class label and that attribute-based classification can not match in terms of performance.

%% file: supplement.tex
\thispagestyle{empty}
\tvspace


\thispagestyle{empty}

In the following supplementary document we provide further details about the proposed work as follows:
\begin{itemize}
    \item Additional experiments with different prompt generation on AWA2 can be found in Section \ref{sec:prompt_awa}.
    \item Additional information about the image used with respect to attribute selection are givben in Section \ref{sec:num_attributes})
    \item In Section \ref{sec:count_occurences}, we provide details about counting class occurrence in Laion400M
    \item A summary of evaluation results in numeric form is listed in Section \ref{sec:detailed_results} for easy comparability.
    \item More qualitative results are shown in Section \ref{sec:qualitative}.
    
\end{itemize}


\begin{figure*}[t]
\centering 
       \begin{subfigure}{1\linewidth}
        
        \begin{subfigure}{0.47\linewidth}
            \centering
            \includegraphics[width=1\linewidth]{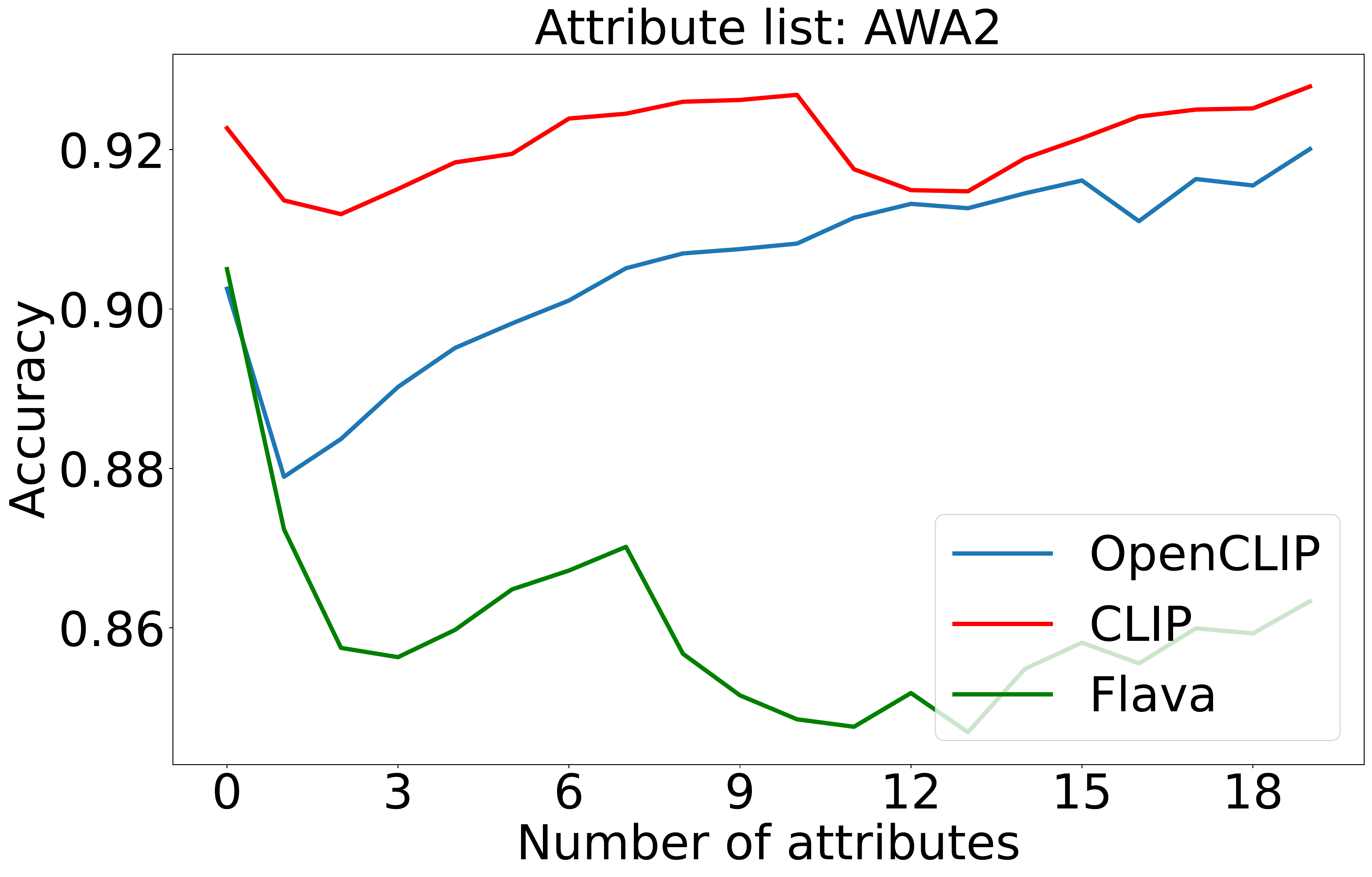}
            \caption{\textbf{CUB-style prompt generation.} An example: ``a photo of a antelope with attributes group, active, walks, longleg, fields, ground, plains''}
        \end{subfigure}\hfill
        \begin{subfigure}{0.47\linewidth}
            \centering
            \includegraphics[width=1\linewidth]{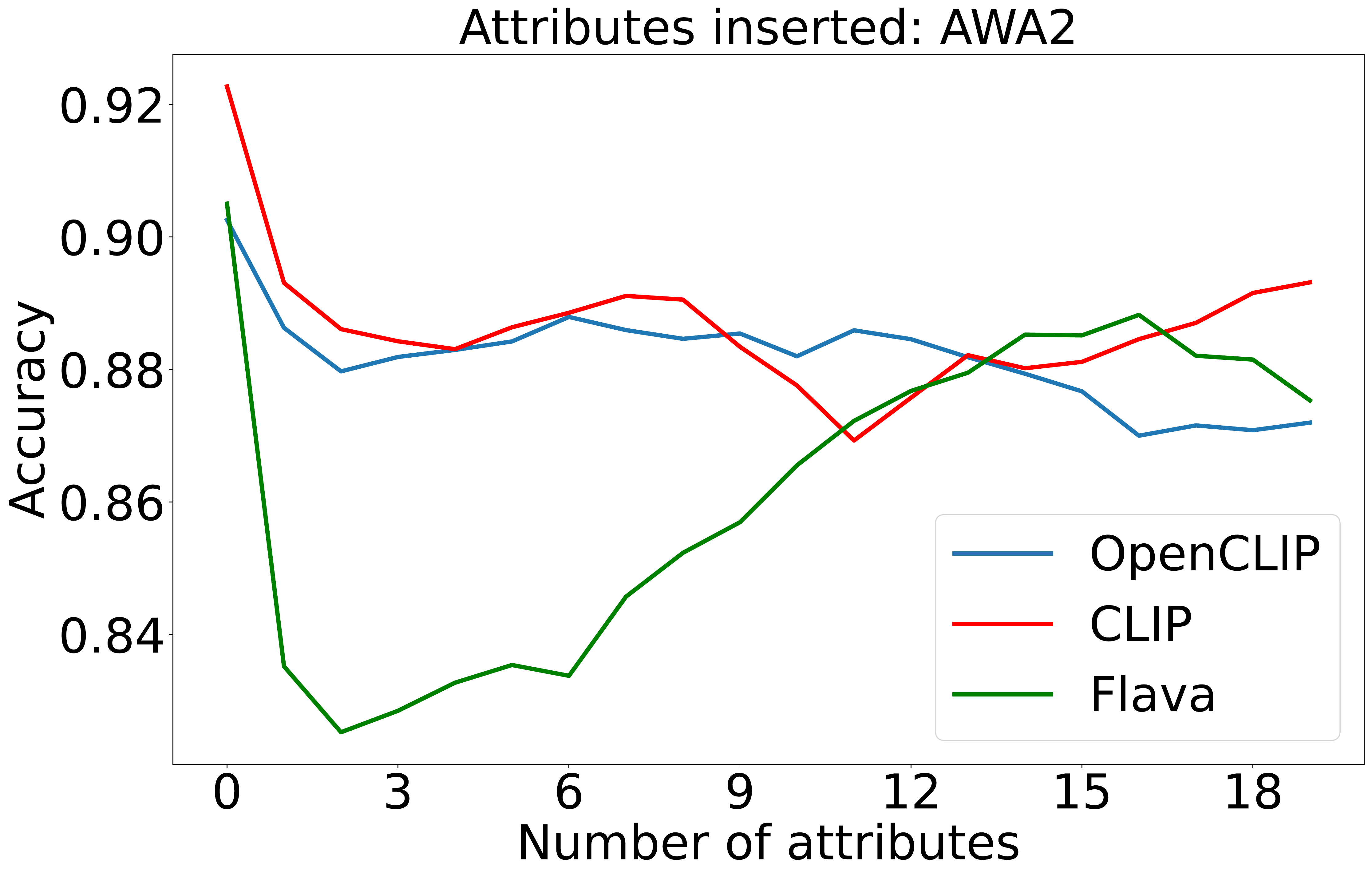}
            \caption{\textbf{Content-based prompt generation.} An example: ``a photo of a group of active antelope that walks that has longleg and is in fields or ground or plains''}
        \end{subfigure}
    \end{subfigure}
  \caption{\textbf{Comparison of evaluation of CLIP, OpenCLIP, FLAVA models on AWA2 and CUB datasets in Setup(3)(Class + class attributes) with (a) CUB-style prompt generation and (b) Content-based prompt generation.} With CUB-style prompt generation, attributes are added at the end of a prompt as a comma-separated list. Instead, with the content-based prompt generation, attributes are inserted into the prompt depending on their part of speech and meaning.
  \label{fig:qual_exp3_new_sent}}
\end{figure*}

\section{Prompt Generation on AWA2} \label{sec:prompt_awa}

The prompts that we use for CUB have a meaningful sentence structure where each attribute is inserted with a description and an expression in a meaningful way, for example ``photo of a woodpecker that has head pattern crested''. 
For AWA2, we added the attributes as a comma-separated list at the end of each prompt, for example ``a photo of a antelope with attributes group, active, walks, longleg, fields, ground, plains''. We did this to ensure that the structure of the prompts for CUB and AWA2 are comparable, and the class label is positioned at the front of each prompt. We refer to this as \textit{CUB-style prompt generation}.

Here, we additionally evaluate if the prompt structure of AWA2 has a significant impact on our findings. Therefore, we create prompts for AWA2 in a different way where we consider the part of speech and \textit{content} of each attribute. For an antelope with seven attributes we create prompts like this: ``a photo of a group of active antelope that walks that has longleg and is in fields or ground or plains''. We refer to this as \textit{content-based prompt generation}.  

In Figure~\ref{fig:qual_exp3_new_sent} we show the evaluation of OpenCLIP, CLIP and FLAVA in \textit{Setup~(3): Class + class attributes} with CUB-style prompt generation and content-based prompt generation. 
In the new prompt structure, attributes are inserted according to their meaning and part of speech. Intuitively, it should be easier for the model to recognize the attributes because they are present in a more natural way. 
However, in this case, the class label is pushed into the middle of the prompt and is not at the beginning of each prompt.
We note that the results for CLIP and Open CLIP with CUB-style prompt generation look similar to the content-based prompt generation, but also that, with content-based prompt generation, accuracy decreases more. 
FLAVA, compared to that, also decreases more with the new prompts but recovers after a few attributes which leads to higher accuracy for FLAVA when many attributes are used, while OpenCLIP and CLIP have a lower accuracy with many added attributes. 
We note that both types of prompts contain the same information (the class label and the same image attributes), and the only difference is the structure of the prompts.


\begin{figure*}[t]
\centering 
    \begin{subfigure}{1\linewidth}
        \begin{subfigure}{0.47\linewidth}
            \includegraphics[width=1\linewidth]{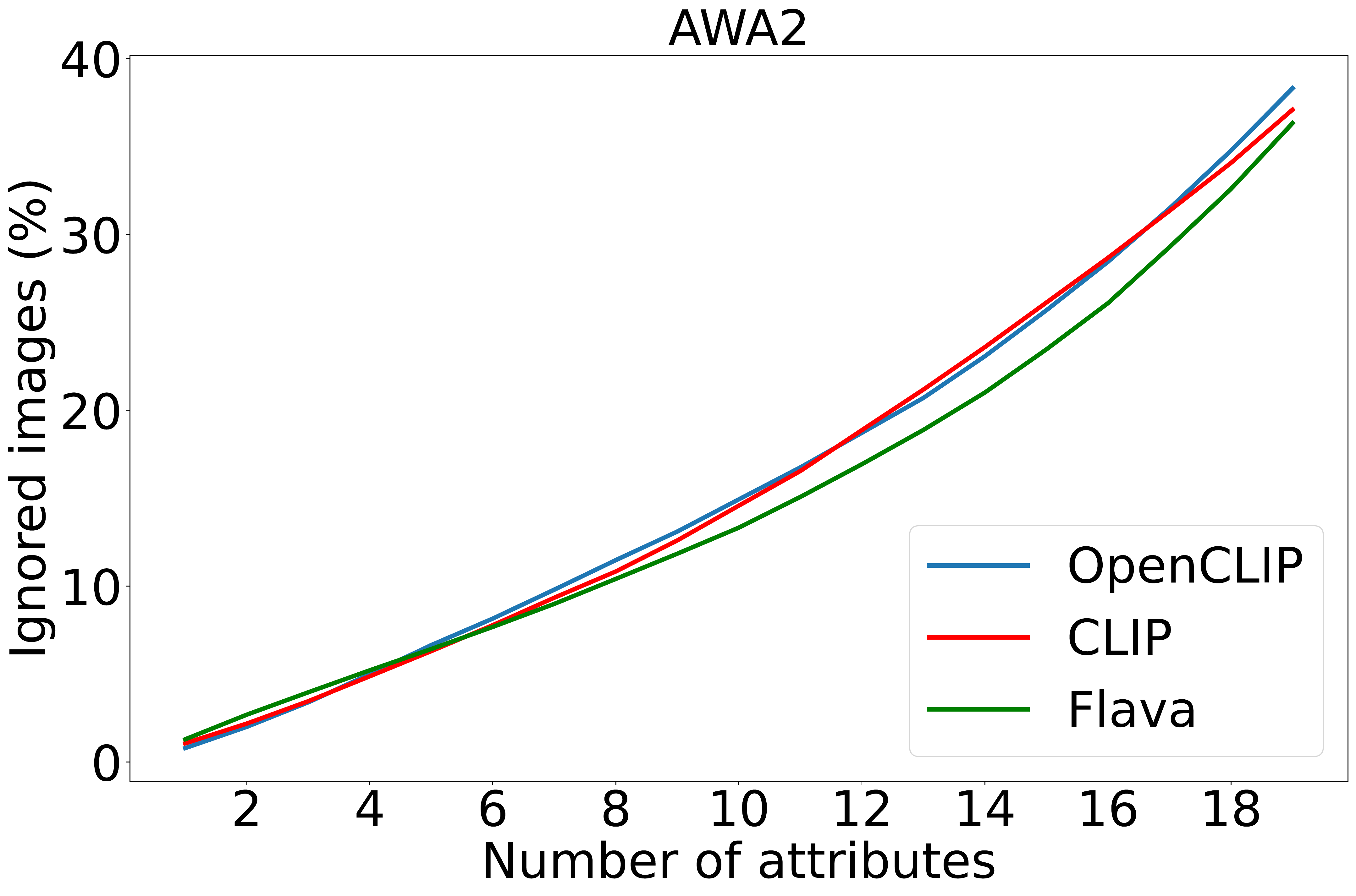}
        \end{subfigure}\hfill
        \begin{subfigure}{0.47\linewidth}
            \includegraphics[width=1\linewidth]{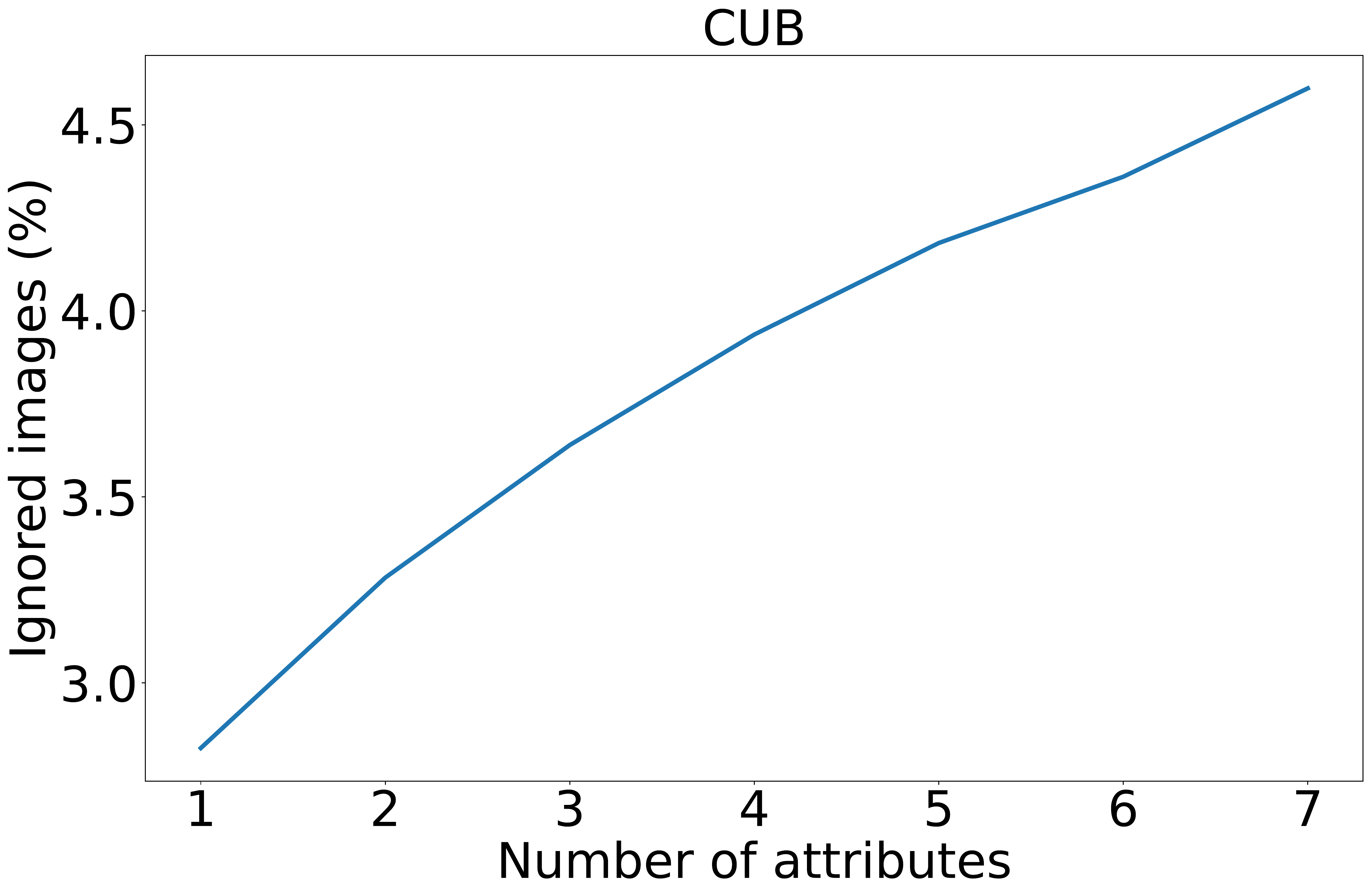}
        \end{subfigure}
    \end{subfigure}
  \caption{\textbf{The average percentage of ignored images in evaluation depending on the number of attributes for OpenCLIP, CLIP, and FLAVA models on AWA2 and CUB datasets.} An image is ignored if the image has fewer attributes than required. On AWA2 image attributes vary from model to model. Therefore, the ignored images are plotted model-wise. For CUB the image attributes don't depend on the model and only one graph for all three models is plotted. Since all experiment Setup(1)--Setup(5) use the same number of image attributes at each step, the ignored images don't vary from experiment to experiment.\label{fig:exp4_awa2_cub1ignored}}
\end{figure*}

\begin{table*}[t]
\centering
\resizebox{\textwidth}{!}{

\begin{tabular}{l| ccccc ccccc ccccc cccc}
\toprule
\multicolumn{1}{c|}{Model} & \multicolumn{19}{c}{Attribute number} \\
\midrule
& 1& 2 &3 &4 &5 &6 &7 &8 &9 &10 &11 &12&13 &14&15 &16 &17 &18 &19 \\
\midrule
OpenCLIP&$0.0079$& $0.02$& $0.0339$& $0.0499$& $0.0665$& $0.0816$& $0.098$& $0.1148$& $0.131$& $0.1493$& $0.1675$& $0.1872$& $0.207$& $0.2307$& $0.2569$& $0.2844$& $0.3148$& $0.3477$& $0.3831$  \\
CLIP&$0.0106$& $0.0218$& $0.0346$& $0.0487$& $0.063$& $0.0778$& $0.0934$& $0.1084$& $0.126$& $0.1458$& $0.1655$& $0.1887$& $0.2118$& $0.2359$& $0.2613$& $0.2867$& $0.3134$& $0.3407$& $0.3709$ \\
FLAVA&$0.0129$& $0.0269$& $0.0396$& $0.0521$& $0.0643$& $0.0768$& $0.0899$& $0.1041$& $0.1185$& $0.1333$& $0.1507$& $0.1693$& $0.1888$& $0.2101$& $0.2347$& $0.2609$& $0.2928$& $0.3259$& $0.3632$   \\
\bottomrule
\end{tabular}
}
\caption{\textbf{Rate of ignored images for OpenCLIP, CLIP, and FLAVA models on the AWA2 dataset.}
Rate of images that are ignored in Setup(1)--Setup(5) (Setup (1): Class + instance attributes;
Setup (2): Class + adversarial instance attributes;
Setup (3): Class + class attributes;
Setup (4): Attributes only;
Setup (5): Adversarial class vs. adversarial attributes.)
\label{table:overviewVLmodelsdata11} }
\end{table*}

\begin{table*}[t]
\centering
\begin{tabular}{l| ccccc cc }
\toprule
\multicolumn{1}{c|}{Model} & \multicolumn{7}{c}{Attribute number} \\
\midrule
& 1& 2 &3 &4 &5 &6 &7 \\
\midrule
OpenCLIP&$0.0282$& $0.0328$& $0.0364$& $0.0394$& $0.0418$& $0.0436$& $0.046$  \\
CLIP&$0.0282$& $0.0328$& $0.0364$& $0.0394$& $0.0418$& $0.0436$& $0.046$ \\
FLAVA&$0.0282$& $0.0328$& $0.0364$& $0.0394$& $0.0418$& $0.0436$& $0.046$   \\
\bottomrule
\end{tabular}
\nvspace
\caption{\textbf{Rate of ignored images for OpenCLIP, CLIP, and FLAVA models on the CUB dataset.}
Rate of images that are ignored in CUB because too few attributes are assigned per image. This rate is the same for all experiment setups (1) - (5) because in each setup the same number of image attributes is used. The ignorance rates are the same for each model because the number of image attributes doesn't depend on the model for CUB.
(Setup (1): Class + instance attributes;
Setup (2): Class + adversarial instance attributes;
Setup (3): Class + class attributes;
Setup (4): Attributes only;
Setup (5): Adversarial class vs. adversarial attributes).
\label{table:overviewVLmodelsdata22} }
\nvspace
\end{table*}

\section{Number of Attributes}  \label{sec:num_attributes}
We report results for Setup(1) -- Setup(5) with a different number of attributes: up to 7 attributes for the CUB dataset, and up to 20 attributes for the AWA2 dataset, but also pointed out that not all images have been annotated with enough attributes.
If fewer attributes are assigned for an image than the step of an experiment setup requires, the image is ignored. 

We therefore assess the rate of ignored images in Figure~\ref{fig:exp4_awa2_cub1ignored}. Since we assigned image attributes on the AWA2 dataset model-wise, the rate of ignored images on AWA2 depends on the model.
Note that the rate of ignored images is independent of the experiment Setup(1) -- Setup(5). 
We further show the exact rates of ignored images for the AWA2 dataset in Table~\ref{table:overviewVLmodelsdata11} and for the CUB dataset in Table~\ref{table:overviewVLmodelsdata22}. We note that for AWA2 the rate of ignored images is higher than for CUB since AWA2 provides only class attributes not image attributes, and the attributes are filtered per image as described in Section ``Attribute detection for AWA2'' of the main paper. This filtering leaves only approximately 20 attributes per image for AWA2, excluding attributes with similarity to an image lower than cut-off values. For OpenCLIP, CLIP, FLAVA the cut-off value is set to $0.2$, $0.24$, $0.245$ which keeps $19.67$, $20.61$, $20.10$ attributes on average per image out of $30.99$ class attributes per class on average. 



\section{Counting Class Occurrence in Laion400M}  \label{sec:count_occurences}

For our analysis of the true zero-shot capabilities of OpenCLIP with respect to the number of class occurrences in the Laion400M dataset (training dataset of OpenCLIP),
we need to count training samples in the Laion400M dataset that correspond to classes of CUB, AWA2, and ImageNet. 
To do this, we search for class labels in the image descriptions of Laion400M samples. We assume that the number of training samples that include the class label in the textual image description is correlated with the true number of training samples that correspond to the class label.

We rely on a direct text search. In order to avoid counting words for the wrong class because of subwords (e.g. "cat" in "catbird"), we add prefixes and suffixes to the class labels.
The prefix is always whitespace. The suffix can be a whitespace or punctuation: ".", "!", "?", ":", " ", ",", ";", "-". To cover plurals "s" or "es" can be added before the suffix.
For each class label, we create all combinations of the prefix, label, plural/no plural, and suffix, which results in 24 different search terms for each class label.
For each dataset, we create a search tree out of all search terms for all class labels according to the Aho Corasick algorithm. Whitespace is added at the beginning and at the end of each text of Laion400M to also find class labels that stand at the beginning or the end of the text and would not have a prefix or suffix otherwise. 
The Laion400M data is processed by the search tree and we obtain the number of samples in Laion400M that correspond to each class label.


\section{Detailed Evaluation Results} \label{sec:detailed_results}
In the main paper, experimental evaluation results for Setup(1)--Setup(5) are plotted in figures. Here we additionally include the exact values of accuracy for all five Setups in Table~\ref{table:overviewVLmodelsdata1} (for the AWA2 dataset) and in Table~\ref{table:overviewVLmodelsdata2} (for the CUB dataset).

\begin{table*}[t]
\centering
\resizebox{1\textwidth}{!}{
\begin{tabular}{l|ccccc ccccc ccccc ccccc c}
\toprule
\multicolumn{1}{c|}{Model} & \multicolumn{20}{c|}{Attribute number} \\
\midrule
&0 & 1& 2 &3 &4 &5 &6 &7 &8 &9 &10 &11 &12&13 &14&15 &16 &17 &18 &19 \\
\midrule
\multicolumn{21}{c}{Setup (1): Class + instance attributes}\\
\midrule
OpenCLIP&$0.9026$& $0.7622$& $0.7772$& $0.7869$& $0.8005$& $0.8079$& $0.8143$& $0.8108$& $0.8118$& $0.8128$& $0.8154$& $0.8216$& $0.8268$& $0.8275$& $0.8322$& $0.8361$& $0.8422$& $0.8462$& $0.852$& $0.8554$  \\
CLIP&$0.9229$& $0.9494$& $0.954$& $0.9548$&$ 0.9555$& $0.9535$& $0.9524$& $0.9504$& $0.947$& $0.9431$& $0.9377$& $0.917$& $0.8896$& $0.8972$& $0.9251$& $0.9329$& $0.9272$& $0.9248$& $0.9165$& $0.9097$ \\
FLAVA&$0.905$& $0.773$& $0.7273$& $0.6723$& $0.5929$& $0.4816$& $0.3924$& $0.3153$& $0.2561$& $0.2286$& $0.2087$& $0.1994$& $0.1895$& $0.18$& $0.1747$& $0.1766$& $0.1818$& $0.1869$& $0.1954$& $0.1997$   \\
\midrule
\multicolumn{21}{c}{Setup (2): Class + adversarial instance attributes}\\
\midrule
OpenCLIP&$0.9026$& $0.8832$& $0.8809$& $0.8835$& $0.8833$& $0.8834$& $0.8827$& $0.8809$& $0.8817$& $0.8805$& $0.8801$& $0.8818$& $0.881$& $0.8819$& $0.8846$& $0.8845$& $0.8861$& $0.8889$&$ 0.8898$& $0.8941$  \\
CLIP&$0.9227$& $0.9118$& $0.9084$& $0.9056$& $0.905$& $0.9037$& $0.9028$& $0.9025$& $0.9033$& $0.9049$& $0.8999$& $0.8981$& $0.8977$& $0.8952$& $0.8987$& $0.9015$& $0.902$& $0.9057$& $0.9092$& $0.9093$ \\
FLAVA&$0.905$& $0.8922$& $0.8902$& $0.8878$& $0.887$& $0.8854$& $0.885$& $0.8829$& $0.8824$& $0.8809$& $0.8818$& $0.8781$& $0.8748$& $0.8729$& $0.8726$& $0.8693$& $0.8681$& $0.8654$& $0.8638$& $0.8651$   \\
\midrule
\multicolumn{21}{c}{Setup (3): Class + class attributes}\\
\midrule
OpenCLIP&$0.9026$& $0.879$& $0.8837$& $0.8902$& $0.8951$& $0.8982$& $0.9011$& $0.9051$& $0.907$& $0.9075$& $0.9082$& $0.9115$& $0.9132$& $0.9127$& $0.9145$& $0.9161$& $0.911$& $0.9163$& $0.9155$& $0.9201$  \\
CLIP&$0.9227$& $0.9136$& $0.9119$& $0.9151$& $0.9184$& $0.9195$& $0.9239$& $0.9245$& $0.926$& $0.9262$& $0.9269$& $0.9175$& $0.9149$& $0.9148$& $0.9189$& $0.9214$& $0.9242$& $0.925$& $0.9252$& $0.9279$ \\
FLAVA&$0.905$& $0.8724$& $0.8575$& $0.8563,$&$0.8597$& $0.8648$& $0.8672$& $0.8702$& $0.8567$& $0.8515$& $0.8485$& $0.8476$& $0.8518$& $0.8469$& $0.8549$& $0.8581$& $0.8555$& $0.8599$& $0.8593$& $0.8633$   \\
\midrule
\multicolumn{21}{c}{Setup (4): Attributes only}\\
\midrule
OpenCLIP&-&$0.0866$& $0.1443$& $0.1878$& $0.2298$& $0.259$& $0.2916$& $0.3197$& $0.3407$& $0.3606$& $0.3819$& $0.4094$& $0.4151$& $0.4285$& $0.4329$& $0.4489$& $0.4542$& $0.4593$& $0.4624$& $0.4675$  \\
CLIP&-&$0.0998$& $0.1523$& $0.1987$& $0.2418$& $0.2746$& $0.3036$& $0.3263$& $0.3515$& $0.3614$& $0.3726$& $0.3454$& $0.3483$& $0.353$& $0.371$& $0.3811$& $0.3835$& $0.3909$& $0.3931$& $0.3977$ \\
FLAVA&-&$0.0683$& $0.1086$& $0.1451$& $0.1757$& $0.2022$& $0.2267$& $0.2443$& $0.239$& $0.241$& $0.246$& $0.247$& $0.266$& $0.2991$& $0.3229$& $0.3358$& $0.3336$& $0.3524$& $0.3622$& $0.3756$   \\
Upper bound (Oracle)&-&$0.0941$& $0.2613$& $0.4478$& $0.6232$& $0.7599$& $0.8517$& $0.9093$& $0.9459$& $0.967$& $0.9794$& $0.9871$& $0.992$& $0.9944$& $0.9964$& $0.9974$& $0.9985$& $0.999$& $0.9993$& $0.9993$\\
\midrule
\multicolumn{21}{c}{Setup (5): Adversarial class vs. adversarial attributes}\\
\midrule
OpenCLIP&-&$[0.9844$& $0.975$& $0.9667$& $0.9625$& $0.959$& $0.9581$& $0.9543$& $0.9529$& $0.9486$& $0.9472$& $0.945$& $0.9456$& $0.9452$& $0.9436$& $0.947$& $0.9484$& $0.9535$& $0.9528$& $0.9551$  \\
CLIP&-&$0.9901$& $0.9845$& $0.9823$& $0.9797$& $0.9808$& $0.9792$& $0.9796$& $0.9814$& $0.9798$& $0.9797$& $0.9788$& $0.9797$& $0.9775$& $0.9784$& $0.981$& $0.9825$& $0.9845$& $0.9878$& $0.9881$ \\
FLAVA&-&$0.9906$& $0.9829$& $0.9776$& $0.9714$& $0.9698$& $0.9692$& $0.9635$& $0.9638$& $0.9599$& $0.9558$& $0.9508$& $0.9492$& $0.946$& $0.9383$& $0.9338$& $0.9284$& $0.924$& $0.9205$& $0.9181$   \\
\bottomrule
\end{tabular}
}
\caption{\textbf{Accuracy for OpenCLIP, CLIP, and FLAVA models for Setup(1) -- Setup(5) on the AWA2 datasets.}
Setup (1): Class + instance attributes;
Setup (2): Class + adversarial instance attributes;
Setup (3): Class + class attributes;
Setup (4): Attributes only;
Setup (5): Adversarial class vs. adversarial attributes.
\label{table:overviewVLmodelsdata1}}
\end{table*}

\begin{table*}[t]
\centering
\begin{tabular}{l| ccccc ccc }
\toprule
\multicolumn{1}{c|}{Model} & \multicolumn{8}{c}{Attribute number} \\
\midrule
& 0 & 1& 2 &3 &4 &5 &6 &7 \\
\midrule
\multicolumn{9}{c}{Setup (1): Class + instance attributes}\\
\midrule
OpenCLIP&$0.6097$& $0.5953$& $0.5887$& $0.5876$&$0.5857$& $40.5892$& $0.5923$& $0.5942$  \\
CLIP&$0.5136$& $0.5655$& $0.623$& $0.6132$& $0.6185$& $0.6126$& $0.6028$& $0.5524$ \\
FLAVA&$0.4382$& $0.3406$& $0.3132$& $0.2712$& $0.2614$& $0.2513$& $0.2598$& $0.2677$   \\
\midrule
\multicolumn{9}{c}{Setup (2): Class + adversarial instance attributes}\\
\midrule
OpenCLIP&$0.6097$& $0.5847$& $0.5757$& $0.5716$& $0.5691$& $0.5701$& $0.5726$& $0.568$  \\
CLIP&$0.5133$& $0.5144$& $0.5121$& $0.5124$& $0.5153$& $0.5177$& $0.5128$& $0.5092$ \\
FLAVA&$0.4382$& $0.441$& $0.4409$& $0.4352$& $0.4216$& $0.4167$& $0.4158$& $0.4081$   \\

\midrule
\multicolumn{9}{c}{Setup (3): Class + class attributes}\\
\midrule
OpenCLIP&$0.6097$& $0.5766$& $0.5792$& $0.574$& $0.5819$& $0.5799$& $0.5918$& $0.5895$  \\
CLIP&$0.5141$& $0.5025$& $0.5093$& $0.5039$& $0.5192$& $0.5211$& $0.5203$& $0.5161$ \\
FLAVA&$0.4382$& $0.4124$& $0.4272$& $0.4308$& $0.4248$& $0.4266$& $0.4273$& $0.4271$   \\
\midrule
\multicolumn{9}{c}{Setup (4): Attributes only}\\
\midrule
OpenCLIP&-&$0.0271$& $0.0358$& $0.0448$& $0.0538$& $0.0572$& $0.0629$& $0.0706$  \\
CLIP&-&$0.0256$& $0.0306$& $0.0343$& $0.0434$& $0.046$& $0.0489$& $0.051$ \\
FLAVA&-&$0.0208$& $0.0253$& $0.0279$& $0.0366$& $0.0413$& $0.0497$& $0.0513$   \\
Upper bound (Oracle)&-&$0.0268$& $0.1142$& $0.3114$& $0.5526$& $0.74$& $0.8518$& $0.9127$\\
\midrule
\multicolumn{9}{c}{Setup (5): Adversarial class vs. adversarial attributes}\\
\midrule
OpenCLIP& -&$0.9775$& $0.977$& $0.9771$& $0.9762$& $0.9747$& $0.9744$& $0.9761$  \\
CLIP&-&$0.9771$& $0.9754$& $0.9783$& $0.9767$& $0.9767$& $0.9762$& $0.9781$ \\
FLAVA&-&$0.9721$& $0.9748$& $0.974$& $0.97$& $0.9687$& $0.9635$& $0.9579$   \\
\bottomrule
\end{tabular}
\caption{\textbf{Accuracy of OpenCLIP, CLIP, and FLAVA models for Setup(1)--Setup(5) on the CUB dataset.}
Setup (1): Class + instance attributes;
Setup (2): Class + adversarial instance attributes;
Setup (3): Class + class attributes;
Setup (4): Attributes only;
Setup (5): Adversarial class vs. adversarial attributes.
\label{table:overviewVLmodelsdata2} }
\end{table*}

\section{More Qualitative Results}  \label{sec:qualitative}

\begin{figure*}[]
   \centering
  \includegraphics[width=0.85\linewidth]{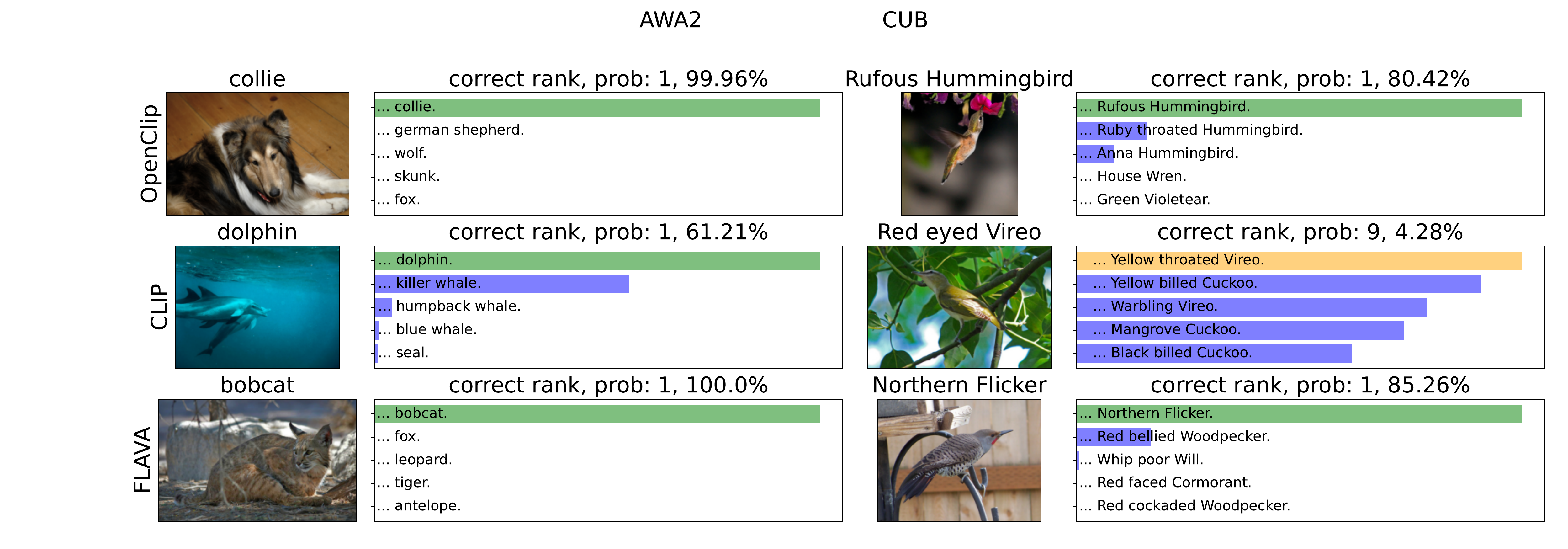}
  \nvspace
  \caption{\textbf{Qualitative examples of predictions of OpenCLIP, CLIP, and FLAVA models on CUB and AWA2 datasets with prompts with no attributes.} 
  Retrieving a text corresponding to an image encoding using different models. The text gallery set is formed by combining a class label (out of all possible classes) with no attribute for each query image. As can be seen in many cases texts corresponding to semantically similar classes are retrieved in the top options.
  \label{fig:qual_exp5}}
  \nvspace
\end{figure*}

\begin{figure*}[]
   \centering
  \includegraphics[width=0.85\linewidth]{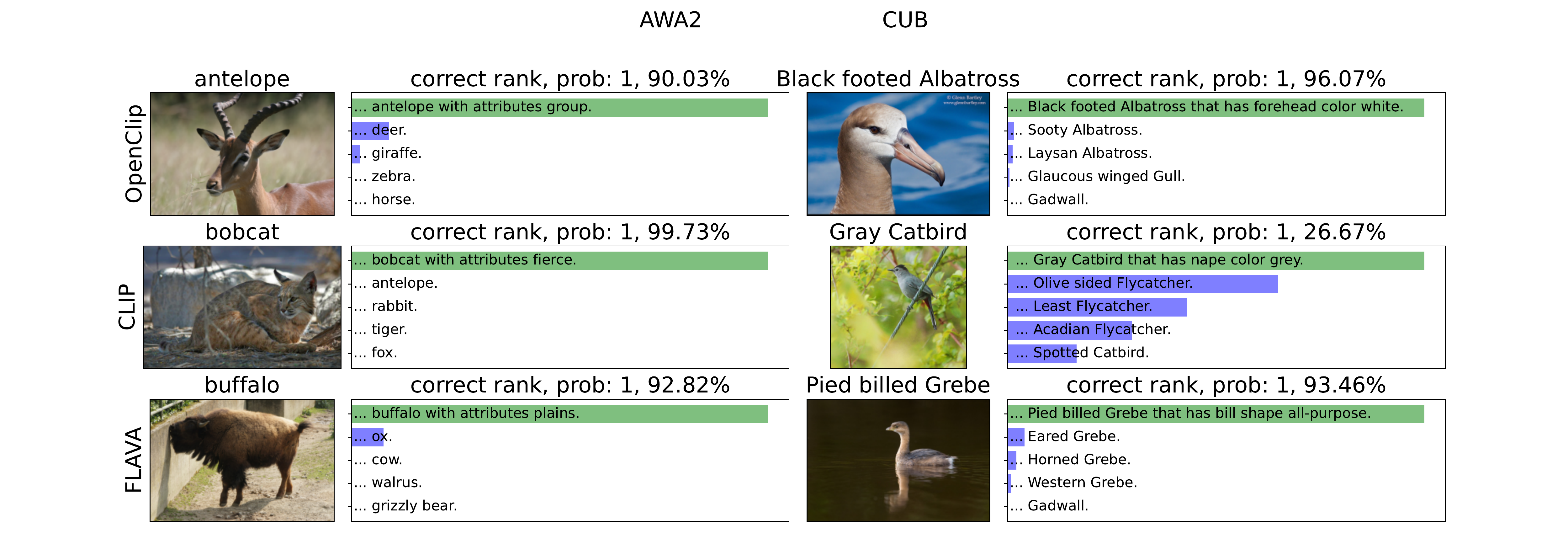}
  \nvspace
  \caption{\textbf{Qualitative examples of predictions of OpenCLIP, CLIP, and FLAVA models on CUB and AWA2 datasets in Setup(1).} 
  Retrieving a text corresponding to an image encoding using different models. The text gallery set is formed by combining a class label (out of all possible classes) with a single image attribute for the image class and only the class label for every other class, according to Setup(1). As can be seen in many cases texts corresponding to semantically similar classes are retrieved in the top options.
  \label{fig:qual_exp1}}
  \nvspace
\end{figure*}

\begin{figure*}[]
   \centering
  \includegraphics[width=0.85\linewidth]{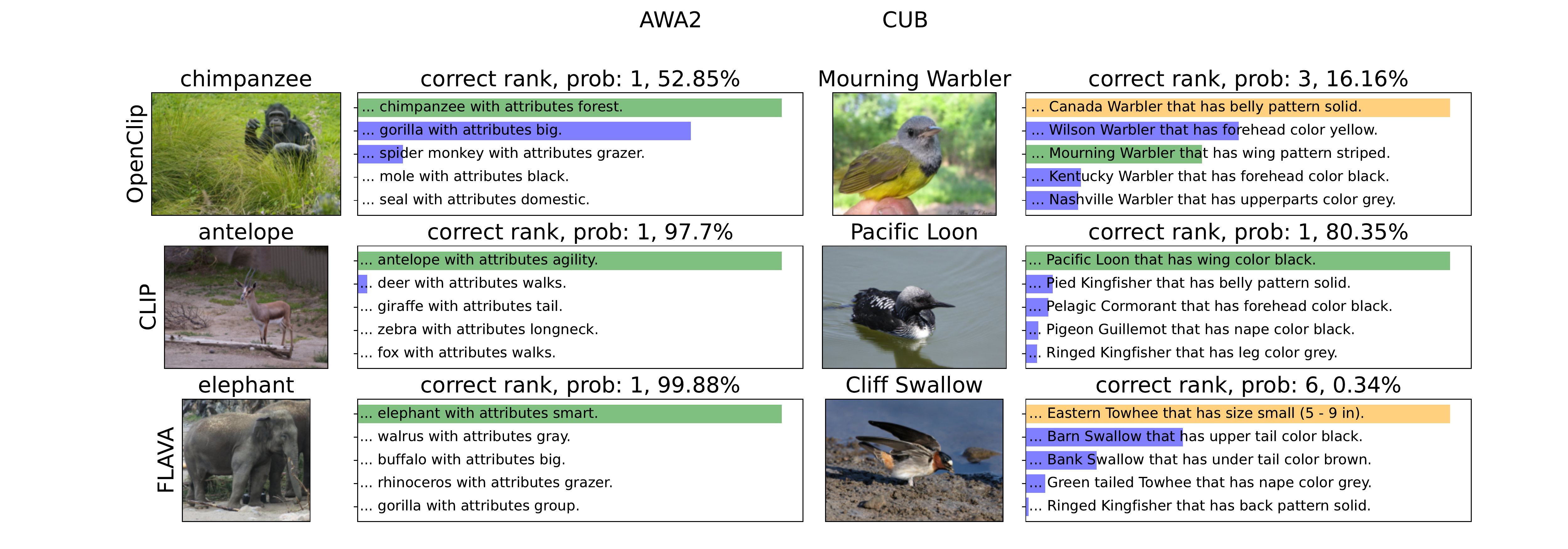}
  \nvspace
  \caption{\textbf{Qualitative examples of predictions of OpenCLIP, CLIP, and FLAVA models on CUB and AWA2 datasets in Setup(3).} 
  Retrieving a text corresponding to an image encoding using different models. The text gallery set is formed by combining a class label with an image attribute for the instance prompt, and the class label and a single class attribute for every other class, according to Setup(3). As can be seen in many cases texts corresponding to semantically similar classes are retrieved in the top options.
  \label{fig:qual_exp3}}
  \nvspace
\end{figure*}

We finally extend our qualitative analysis by presenting qualitative examples of model predictions in \textit{Setup~(1) (Class + instance attributes)} in Figure~\ref{fig:qual_exp1}, in \textit{Setup~(3) (Class + class attributes)} in Figure~\ref{fig:qual_exp3}, and model prediction without any attributes given in Figure~\ref{fig:qual_exp5}.

We note, that even without attributes given (Figure~\ref{fig:qual_exp5}), all models have high  top-1 accuracy, but also tend to predict semantically related classes among top-5 results. 
Such as for an image of a  \textit{collie}, only dog species are in the top-5 predictions. For a \textit{dolphin}, only sea animals are in the top-5 predictions. And for a \textit{bobcat}, where the model is extremely confident, nevertheless, mainly animals with a similar physical appearance are listed in the top-5 predictions.
When we take a closer look at the predictions for the CUB dataset, we observe that mostly similar bird species are predicted in the top-5 predictions. For example several subspecies of the same species (as for \textit{rufous hummingbird}) or bird species that have a similar physical appearance. 

A similar observation can be made in Figure~\ref{fig:qual_exp1} that corresponds to experiment Setup (1) (Class + instance attributes) where a prompt with image attributes and class label is compared to prompts with only the class label, and in Figure~\ref{fig:qual_exp3} that corresponds to experimental Setup (3) (Class + class attributes) where a prompt with image attributes and class label is compared to prompts with class attributes and class label. Independently of the attribute configuration in the prompts, the models seem to behave similarly.

